\PassOptionsToPackage{square,comma,numbers,sort&compress}{natbib}

\documentclass[final,5p,times,twocolumn]{elsarticle}

\usepackage{amssymb}
\usepackage{svg}
\usepackage{siunitx}
\usepackage{graphicx}

\usepackage{lineno}
\usepackage{caption}
\usepackage{subcaption}

\usepackage{xcolor}
\usepackage{makecell}

\journal{Energy and Buildings}

\begin{document}



\title{Longitudinal thermal imaging for scalable non-residential HVAC and occupant behaviour characterization}

\author{Vasantha Ramani$^{1}$, Miguel Martin$^{1}$, Pandarasamy Arjunan$^{1}$, Adrian Chong$^{2}$, Kameshwar Poolla$^{3}$, Clayton Miller$^{2, *}$}

\address{$^{1}$Berkeley Education Alliance for Research in Singapore, Singapore}
\address{$^{2}$College of Design and Engineering, National University of Singapore (NUS), Singapore}
\address{$^{3}$Department of Electrical Engineering and Computer Sciences, University of California, Berkeley, CA, USA}

\address{$^*$Corresponding Author: clayton@nus.edus.sg, +65 81602452}



            


\begin{abstract}
This work presents a study on the characterization of the air-conditioning (AC) usage pattern of non-residential buildings from longitudinal thermal images collected at the urban scale. The operational pattern of two different air-conditioning systems (water-cooled systems operating on a pre-set schedule and window AC units operated by the occupants) are studied from the thermal images. It is observed that for the water-cooled system, the difference between the rate of change of the window and wall temperature can be used to extract the operational pattern.  While, in the case of the window AC units, wavelet transform of the AC unit temperature is used to extract the frequency and time domain information of the AC unit operation. The results of the analysis are compared against the indoor temperature sensors installed in the office spaces of the building. This forms one of the first few studies on the operational behavior of HVAC systems for non-residential buildings using the longitudinal thermal imaging technique. The output from this study can be used to better understand the operational and occupant behavior, without requiring to deploy a large array of sensors in the building space.
\end{abstract}



\maketitle

\section{Introduction}
The energy sector has been one of the major contributors to greenhouse gas (GHG) emissions and climate change. Electricity and heating are one of the largest sub-sector that contributes close to 24\% of the GHG emissions \cite{lamb2021review}. Energy consumption is expected to increase in the future due to the growing demand for heating and cooling. Improvement in building energy performance to reduce energy consumption has been studied extensively and implemented through various technological innovations in the form of energy-efficient building materials, façade systems, high-efficiency heating, ventilation, and air conditioning systems (HVAC), energy-efficient appliances, and others (\cite{ruparathna2016, pacheco2017, belussi2019}). Besides the adoption of these technologies, increased energy savings are also achieved by the optimal operation of the energy systems. The operation of the system is aimed not only to minimize energy usage and cost but also aims at maximizing occupant comfort, well-being, and satisfaction. 

Various approaches such as regular energy audits (\cite{krarti2020energy}), numerical simulation of the built environment (\cite{wang2016advances, harish2016review}), and data-driven approaches involving analyses of the energy meter data (\cite{molina2017data, amasyali2018review}) are adopted for analysis of the building performance. Energy audits and surveys are conducted once every few years to understand the building's energy usage and performance. The amount of metadata that is to be collected depends on the level of energy audit (\cite{ASHRAE211}). A comprehensive energy audit requires information on the environment profile, occupant behaviour, parameters and the number of systems using energy in addition to the energy use data (\cite{ruparathna2016}), which can make the auditing process laborious and time-consuming. On the other hand, numerical simulation of building energy performance typically requires calibrating the model to ensure its reliability (\cite{ashrae2014guideline, evo2014uncertainty}). However, model calibration remains challenging due to a lack of clear guidelines and best practices (\cite{chong2021calibrating}).  Also, the numerical simulation require the knowledge of indoor and outdoor conditions, that are provided as input to the model and accurate estimation of these parameters can be difficult. Data-driven approaches have gained traction in the recent few years because of the availability of smart meter data and other building sensor measurement data. These methods have found applications in energy management such as load profiling (\cite{park2020good, zhu2019data}), anomaly detection \cite{himeur2021artificial}, demand and response studies (\cite{qi2020smart}), and energy forecasting (\cite{amasyali2018review, bourdeau2019modeling}). 

Studying building operational behaviour can assist in understanding the energy usage pattern and also in identifying efficiencies and inefficiencies arising out of the usage pattern. According to the International Energy Agency (\cite{IEA2021cooling}) report on cooling, space cooling accounts for nearly 16\% of building sector final electricity consumption in 2020. Typically, in the case of a commercial building, the demand for cooling is met by a centralized HVAC system. These air conditioning systems are generally operated on a pre-defined schedule, and their operation is targeted at providing comfortable working indoor temperature and humidity conditions for the occupants. This temperature may be different from the outdoor temperature depending on the location of the building. Several research studies have shown that is it possible to determine the duration of operation of the air conditioning system using the measured indoor temperature. In the case of residential buildings, the AC operation largely depends on the occupant behaviour, weather conditions, socio-economic conditions and others (\cite{yasue2013modeling, xia2019study, aqilah2021analysis}). Unlike residential buildings, there are very few research studies conducted for commercial/office buildings (\cite{zhou2021comparison, yun2011field}). In either case, such studies on air conditioning usage patterns have been carried out through instrumentation of the buildings with an extensive network of sensors. 

In the recent work by \cite{arjunan2021operational} possibility to observe the split AC operational behaviour for residential buildings using infrared imaging is demonstrated. Infrared imaging offers a non-contact way of monitoring the operation and does not require extensive sensor instrumentation of the indoor spaces. However, most of the work demonstrates AC usage during the night time. While, in the case of commercial buildings, typical air-conditioning peak cooling loads are observed during the day. Also, in non-residential buildings, cooling of indoor spaces can be achieved using the centralized air-conditioning system or split/window AC units, or a combination of two. In summary, the AC operation in commercial buildings can be quite different from that of residential buildings (\cite{meng2019questionnaire}).

This research aims at providing an alternative means for identifying the operational behaviour of the air conditioning system in an educational building using the longitudinal thermal images. The main research question that is addressed in this study is: ‘How to describe the building HVAC operation using the surface temperature of the buildings measured using a thermal camera at an urban scale?’. The research question is further broken down into sub-questions as follows:

\begin{itemize}
    \item How is the change in surface temperature of windows and walls of the building correlated to air-conditioning operation pattern?
    \item Can the change in the surface temperature of the window AC condenser units be used to identify AC usage patterns?    
    \item What are the conditions for capturing IR images that can yield maximum information gain for this application?
    
\end{itemize}   

To achieve the mentioned objectives, a thermal observatory was installed on a rooftop of a building overseeing some of the educational buildings on the campus of the National University of Singapore. This work is an outcome of the study conducted over a duration of four months. In the following section, some of the studies related to thermal imaging and time series analysis are discussed. The methodology adopted for this study is presented in Section 3. In Section 4, the results from the analysis of thermal images are explained in detail. Finally, conclusions from this study are summarised in Section 5.

\section{Related studies}
\subsection{Infrared (IR) imaging}
Every object emits infrared (IR) radiation (as long as its temperature is greater than absolute zero) depending on its temperature. The IR camera is equipped with suitable sensors to capture the infrared radiation from the objects and creates a thermal visualization of the scene. Observing the changes and differences in temperature profile of objects through thermal imaging has been useful in several applications such as monitoring heat flows, performing quality checks for identifying heating elements in a system, detecting leakages, building inspections and many others (\cite{mollmann2017infrared}). 

One of the main advantage of using IR imaging is that it offers a non-contact technique for scanning a large area in a short time period. A thermal camera consists of an optical system that focuses the radiation from the scene onto a detector called a microbolometer. When the long-wave infrared (LWIR) radiation strikes the detector, it results in changes in its resistance, which is converted to apparent temperature ($T_{obj}$) values estimated as follows (\cite{horny2003fpa}):

\begin{equation}
\label{eq: eq1}
T_{obj} = \frac{B}{ln(\frac{R_1}{R_2(U_tot+O)}+f)}
\end{equation}
where, $U_{tot}$ is the signal response to the LWIR incident on the detector, \textit{B}, $R_{1}$, $R_{2}$, \textit{O} and \textit{F} are the camera calibration constants, which are calibrated based upon the type of camera and application, and are included in the metadata of each thermal image.

The camera output signal ($U_{tot}$) is a result of radiation from three sources which are radiation from the object, radiation reflected by the object, and the radiation transmitted from the atmosphere and is given as follows:

\begin{equation}
\label{eq: eq2}
U_{tot} = \epsilon\tau U_{obj} + \tau(1-\epsilon)U_{refl}+(1-\tau)U_{atm}
\end{equation}
where, $\epsilon$ is the emissivity of the object, $\tau$ is the atmospheric transmissivity, $U_{obj}$ is the radiation in terms of signal response from the object, $U_{refl}$ is the radiation reflected by the object, and $U_{atm}$ is the radiation transmitted from the atmosphere. 

IR imaging have been used extensively in the built environment for carrying out building energy audits (\cite{balaras2002infrared, lucchi2018applications}), detection of structural defects in building (\cite{milovanovic2016review, khan2015multi}), estimating the thermal transmittance (\cite{albatici2015comprehensive}), predicting occupant thermal comfort preferences (\cite{ghahramani2018towards, li2019robust}), and detecting window opening state \cite{fan2021winset}. The temporal and spatial scale at which the imaging is performed can greatly vary depending on the nature of application, and availability of resources. At the spatial level, imaging can be classified as (\cite{Martin2022}): 1) micro-scale, 2) local scale, and 3) mesoscale. Micro-scale studies require a handheld IR camera or Unmanned Ariel Vehicle (UAV) for imaging. In thermal inspection using UAVs a drone with an IR camera is used for conducting building energy audits, retrieving material properties (U-value) of building elements, and structural defect detection. However, such methods require drone flight path planning, and this can increase the cost of the entire operation (\cite{tejedor2022application, tejedor2022non}). Further, it is difficult to obtain longitudinal data using UAV, which is required for describing hourly and diurnal changes in the operation. Mesoscale IR imaging involves the use of IR data captured using satellites and is mainly employed for urban heat island  analysis at a city scale. The local scale is the scale between the two and mainly consist of an observatory for scanning over a distance of a few hundred meters. Studies have shown that Mesoscale IR scanning can be used for  estimation the heat fluxes (\cite{richters2009analysis, sham2012verification}), identifying heat sources in the urban canopy (\cite{dobler2021urban}) and extraction of AC usage pattern in residential units (\cite{arjunan2021operational}). In this work, imaging is performed at the local scale to understand the operation pattern of HVAC system in an educational building.

\subsection{Pattern detection from time series}
Analysis of time series to identify patterns is one of the well researched area (\cite{esling2012time, fu2011review}). Depending on the time-series pattern, insights on physical phenomena can be extracted either from time-domain or frequency domain analysis or both. Fast Fourier Transform (FFT) is commonly used to analyse the time-series in  frequency domain. This has been widely used for various applications such as fault detection, image processing, signal processing, condition monitoring and others. Even though it is possible to obtain the frequency content using FFT, it does not provide any temporal information. Short term Fourier transform is a type of Fourier transform used to determine the frequency content of short segments of time series with time. One of the drawbacks with short term Fourier transform is that the resolution is fixed, which implies the window selected for transform will affect the time or frequency resolution. Using a narrow window can result in poor frequency resolution but better time resolution. While, using a broad window will result in poor time resolution but with better frequency resolution. 

Wavelet transform (\cite{pathak2009wavelet, debnath2002wavelet}), on the other hand overcomes the shortcomings of fixed resolution faced in the case of short term Fourier transform. A Wavelet in a wavelet transform is a wave-like oscillation with a finite duration and zero mean. There are several types of wavelets such as the Morlet, Mexican Hat, Coiflets, Daubechies and others. The choice of wavelet will depend on the type of application. The two aspect of transform is scaling and shifting. Scaling refers to extent of stretching or shrinking of the wavelet and is inversely proportional to the wavelet frequency. While shifting refers to the delay or advancing the wavelet along the duration of signal. Depending on how the wavelet are scaled and shifted the wavelet transform is classified as continuous wavelet transform (CWT) and discrete wavelet transform (DWT). CWT is mainly used for time frequency analysis. While, DWT is used for de-noising and compressing images or signal. In this work, CWT is used for performing time-frequency analysis. CWT of a signal \emph{x}(\emph{t}) expressed as follows:
\begin{equation}
\label{eq: eq3}
X_{w}(a,b) = \frac{1}{\sqrt(|a|)}\int_{-\infty}^{\infty}x(t)\overline{\psi}(\frac{t-b}{a})dx
\end{equation}
where \emph{a} is the scale factor, \emph{b} is the transnational value, and \emph{$\psi$}(\emph{t}) is the mother wavelet that is continuous in both time and frequency domain, and the over line represent the conjugate complex operation.

\section{Methodology}
In this section the methodology adopted for sensor measurements involving urban scale IR imaging, point measurements and data analysis of the thermal data are discussed in detail. 

\subsection{Measurements}
\subsubsection{Urban scale Infrared (IR) observatory}
\begin{figure*}
    \centering
    \includegraphics[width = 375pt]{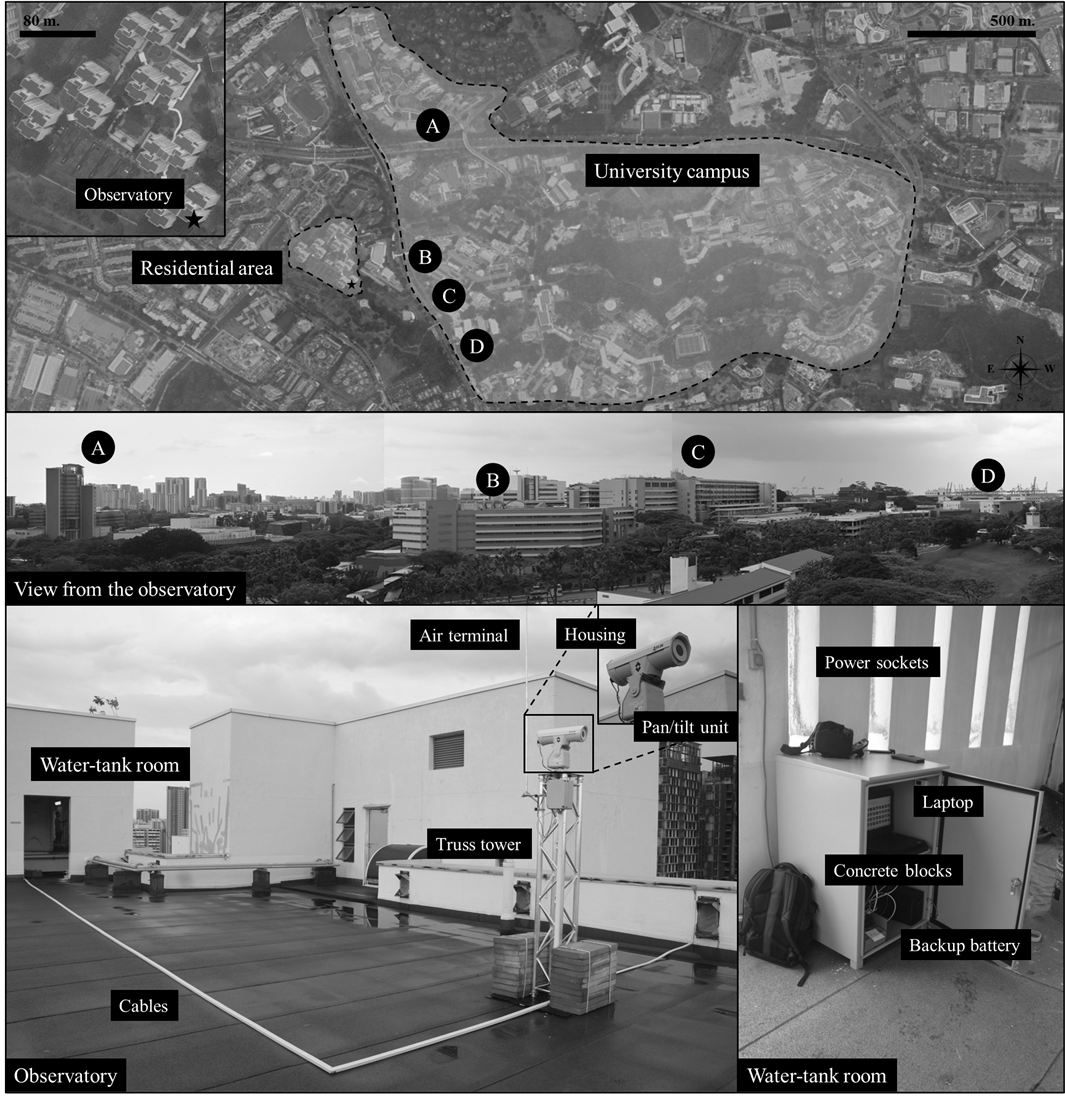}
    \caption{Figure showing the location of the four buildings captured using the thermal, and the IR camera observatory (adapted from \cite{miguel2022b}).}
    \label{figure:IR_observatory}
\end{figure*}

Figure~\ref{figure:IR_observatory} (top) shows the map of the location of the installation of the urban-scale IR observatory. The observatory is installed on the rooftop of a residential building at a height of about 42m from the ground level. To the west of the observatory are the four buildings marked as `A', `B', `C', and `D', and the corresponding digital images as viewed from the observatory are shown in Figure ~\ref{figure:IR_observatory} (center). Building `A' is a fully glazed building, buildings `B' and `C' are concrete buildings with windows and building `D' is a net-zero building. IR observatory installed at the rooftop of the building overseeing these buildings is shown in Figure~\ref{figure:IR_observatory} (bottom left). 

The thermal camera (FLIR 300) is housed in a casing to protect it from the external environment and extreme weather conditions. The specifications and the default Plank's constant of the thermal camera are listed in Table~\ref{tab:Tb_camera_specs} and Table~\ref{tab:Plank'constant} respectively. The camera housed in the casing is mounted on a pan-tilt unit that can rotate along \ang{360} in the horizontal axis. The pan-tilt unit and the camera are installed on a truss tower fixed to the base plate. The concrete blocks are placed on either side of the base plate, which prevents motion/sway of the truss tower due to external loads such as wind. In addition, an air terminal is also installed to protect the observatory from lightning. The pan-tilt unit and the thermal camera are connected to a laptop housed in weather-protected casing in the water tank room located next to the observatory as shown in Figure~\ref{figure:IR_observatory} (bottom right). The pan-tilt unit is programmed to stop at four instances for fifteen seconds each during one rotation cycle in to capture thermal images of four buildings. A software interface developed by NAX Instruments Pte Ltd is used for capturing and storing images in jpg format. Of the four buildings shown in Figure~\ref{figure:IR_observatory}, building `B' and building 'C' are of interest in this study. Indoor spaces of building `B' and most of building `C' are air-conditioned using a water-cooled centralized HVAC system. While some of the office spaces in Building `C' are air-conditioned using window or split AC units. The windows in Building `B' and Building `C' are double glazed and single pane windows respectively.

\begin{table}[]
    \centering
    \begin{tabular}{c|c}
    \hline
    Resolution & 16 bit, 320x240 pixels\\
    Thermal sensitivity & 50mK $@$ 30$^o$C\\
    Sensor & Uncooled Microbolometer FPA\\ 
    Spectral range & 7.5 to 13 $\mu$m\\ 
    Field of view (FOV) & 25$^o$ (H) and 18.8$^o$ (V)\\
    Accuracy & $\pm$2$^o$ or 2\% of reading\\
    Power supply & 110/220 V AC\\
    Weight & 0.7 kg\\
    Size & 170mm x 70mm x 70mm\\
    \hline
    \end{tabular}
    \caption{FLIR A300 thermal camera specification}
    \label{tab:Tb_camera_specs}
\end{table}

\begin{table}[]
    \centering
    \begin{tabular}{c|c}
    \hline   
    \emph{R\textsubscript{1}} & 14911.1846\\
    \emph{R\textsubscript{2}} & 0.0108\\
    \emph{f} & 1.0\\ 
    \emph{O} & -6303.0\\ 
    \emph{B} & 1396.6\\
    \hline
    \end{tabular}
    \caption{FLIR A300 thermal camera default Plank's constant}
    \label{tab:Plank'constant}
\end{table}

\subsubsection{Point temperature measurement}
\begin{figure*}
     \centering
     \begin{subfigure}[b]{0.6\textwidth}
         \centering
         \includegraphics[width=\textwidth]{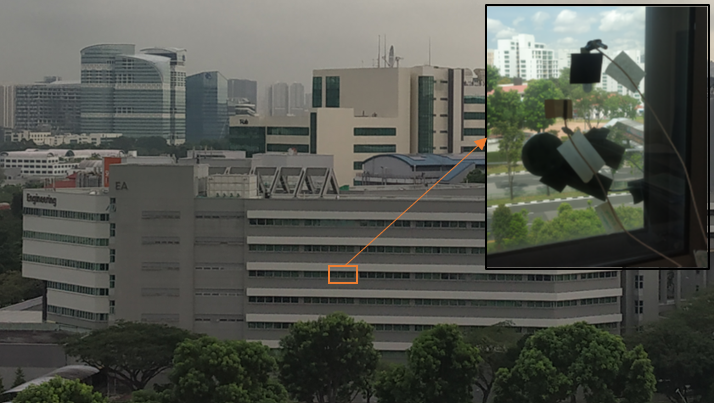}
         \caption{Surface temperature measurement}
     \end{subfigure}
     \begin{subfigure}[b]{0.6\textwidth}
         \centering
         \includegraphics[width=\textwidth]{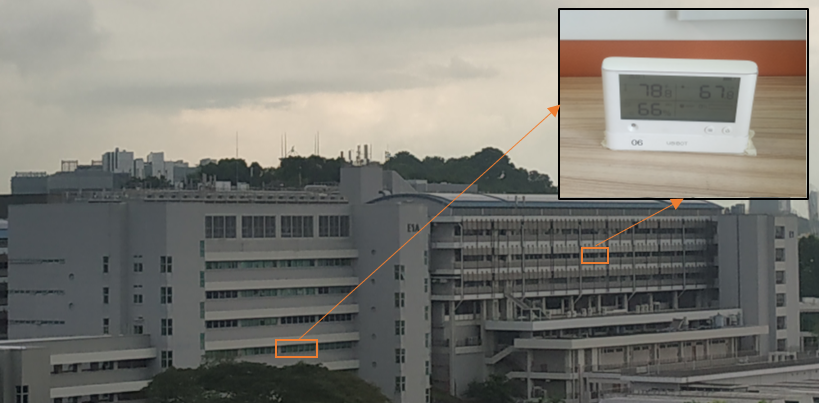}
         \caption{Dry-bulb air temperature}
     \end{subfigure}
        \caption{Digital image of the buildings showing the location of (a) the surface temperature contact sensor, and (b) the dry-bulb indoor air temperature sensor (the red boxes indicates the location of the sensor)}
        \label{figure:Ubi_location}
\end{figure*}

In addition to the surface temperature measured using the thermal camera, indoor dry-bulb air temperature and outdoor surface temperature measurement sensors were installed in some of the office spaces of the buildings. The main purpose for installation of point temperature measurement sensors is to compare the temperature data from the IR image with the ground truth. For monitoring the indoor dry-bulb temperature, UbiBot WS1 pro indoor monitoring sensors (temperature accuracy of ${\pm}$ 0.3$^{\circ}$C) were installed in the indoor spaces next to the windows as shown in Figure~\ref{figure:Ubi_location}. In addition to the indoor temperature sensor, one of the windows of Building `B' was instrumented to monitor its surface temperature. These windows were previously single pane, which was retrofitted with installation of another window pane next to the existing one. Two resistance temperature detector (RTD) sensors were installed on the window surface, one on the exterior surface exposed to the outdoor environment, and another on the window exposed to the indoor environment. These sensors were attached to the surface using thermal conductive putty for good transfer of heat from the surface to the sensor. In addition to the two temperature sensors for measuring the surface temperature, another RTD sensor was installed to monitor the indoor dry-bulb temperature.  

\subsection{Image processing and Time series analysis}
The first step in the analysis of the images is the cleaning of the data. During rain, due to the thermal radiation from the rain droplets, the captured images are not suitable for analysis. Also, it is required to remove the images that are captured during the motion of the pan-tilt unit. To achieve this a classification model using convolution neural network (CNN) (\cite{albawi2017understanding}) is implemented and the images are classified based on the building type and corrupted images (those that are not suitable for analysis). The CNN model consists of three convolution blocks with a max pooling layer for each of them, which is followed by a fully connected layer. The model is trained on 4316 pre-classified images and subsequently used for classification of the entire infrared image dataset. 

The next step after the classification of the images is the extraction of the radiometric data. For this, `flirextractor' python package (\cite{klink2019aloisklink}) is used. The raw values from the image are converted to temperature using Equation \ref{eq: eq1}. The temperature values are stored in comma-separated value (CSV) format, which allows easy and faster access to the temperature data. Subsequently, temperature time series from regions of interest (RoI) in the image are extracted. In this work, the RoI is a set of pixels in the image that corresponds to either wall, window, or AC units. The RoI in the thermal image is identified and segmented using labelme (\cite{russell2008labelme}), an open-source annotation tool. From the thermal data, the temperature of RoI is extracted and stored in CSV format, which can be easily used for further analysis. 

Depending on the nature of the time series and the level of information to be extracted, analysis can be performed in the time-domain or frequency domain, or both. In this work, the operation pattern of two different types of HVAC systems are analysed. The first one is the water-cooled HVAC system that has a pre-decided operation pattern. That is, almost every day the system is switched on and off at a fixed time. The second type is air-cooled window and split AC units, which are operated according to the needs of the occupant. Thus, these two systems require two different methodologies for the analysis of the operation patterns from the thermal time series data extracted from the IR images. The analysis of the operation of HVAC system with pre-defined operation pattern is performed in the time domain. While the analysis of the AC units with separate condenser units is performed mainly in the frequency domain using both Fast Fourier transform and wavelet transform. In the subsequent sections, the analysis and the results from the analysis of operation of the HVAC systems are explained in detail.

\section{Results and analysis}
In this section results from the infrared imaging study to characterize the operational behaviour of the HVAC system are presented.

\subsection{IRIS Dataset}
\begin{figure*}
    \centering
    \includegraphics[width = 350pt]{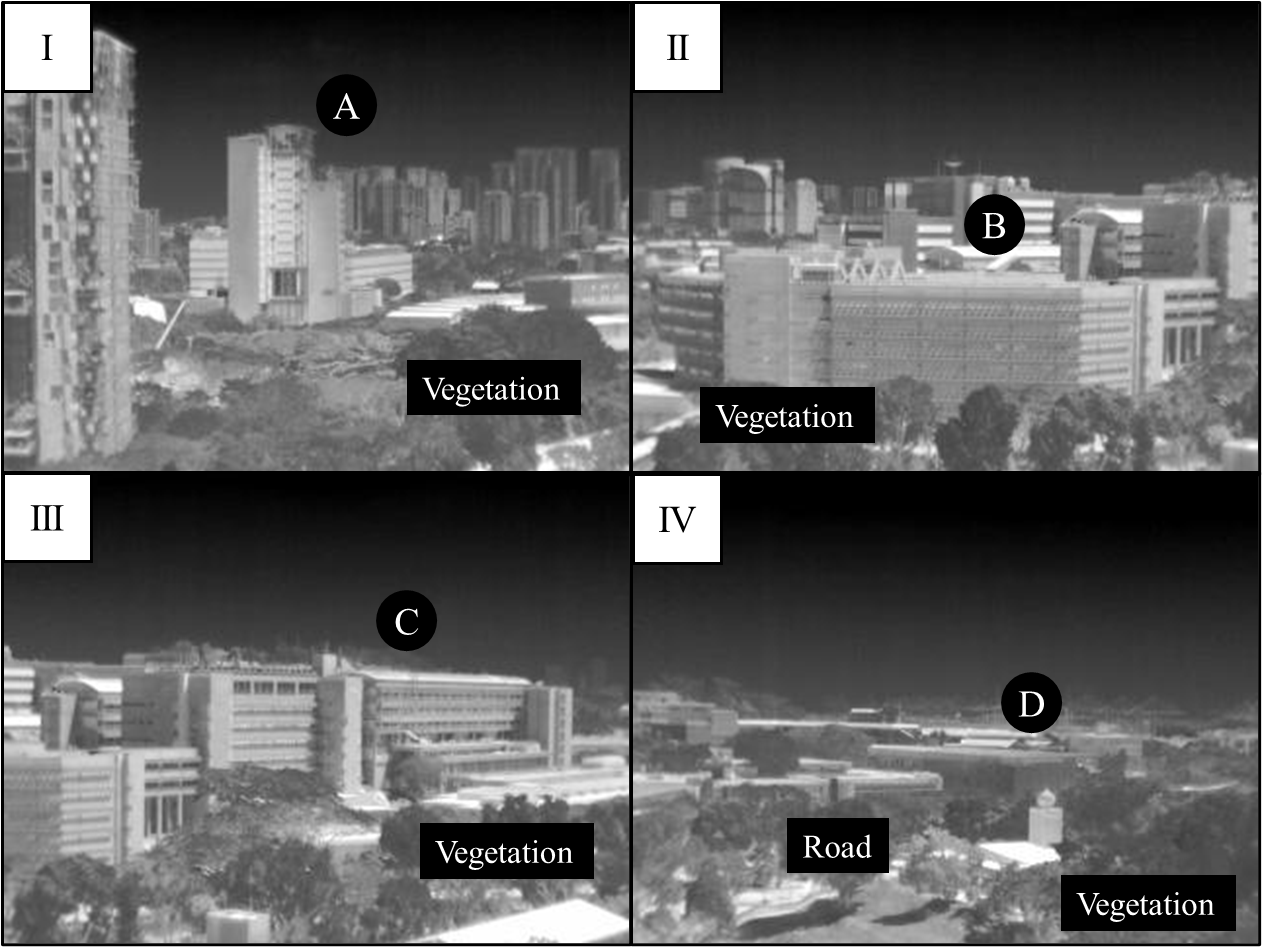}
    \caption{ IR images of the buildings captured using the thermal camera (adapted from \cite{miguel2022b}).}
    \label{figure:Ir_images}
\end{figure*}

Figure~\ref{figure:Ir_images} shows the IR images of the four buildings captured using the thermal camera. The colour contrast helps in identifying the hotter and cooler regions. That is, in the figure shown, the temperature of the bright regions is higher than the darker regions. As described in Section 3.2, a CNN model for classification is developed for the removal of unwanted images and also for classifying the four different buildings. The trained model has an accuracy of 0.98\% on the test dataset and is subsequently used for the classification of the rest of the IR dataset. The urban scale infrared observatory is operated for four months period yielding an infrared image dataset that is rich longitudinally as well as spatially. One of the important aspect of any measurement is its level of accuracy. To verify the surface temperature measurement from the IR camera, the temperature time series data from the IR images are compared against ground truth measurements. This is described in the detail in the subsequent section.

\subsection{Comparison of the extracted IR temperature against ground truth measurement}
\begin{figure}
     \centering
     \begin{subfigure}[b]{0.5\textwidth}
         \centering
         \includegraphics[width=\textwidth]{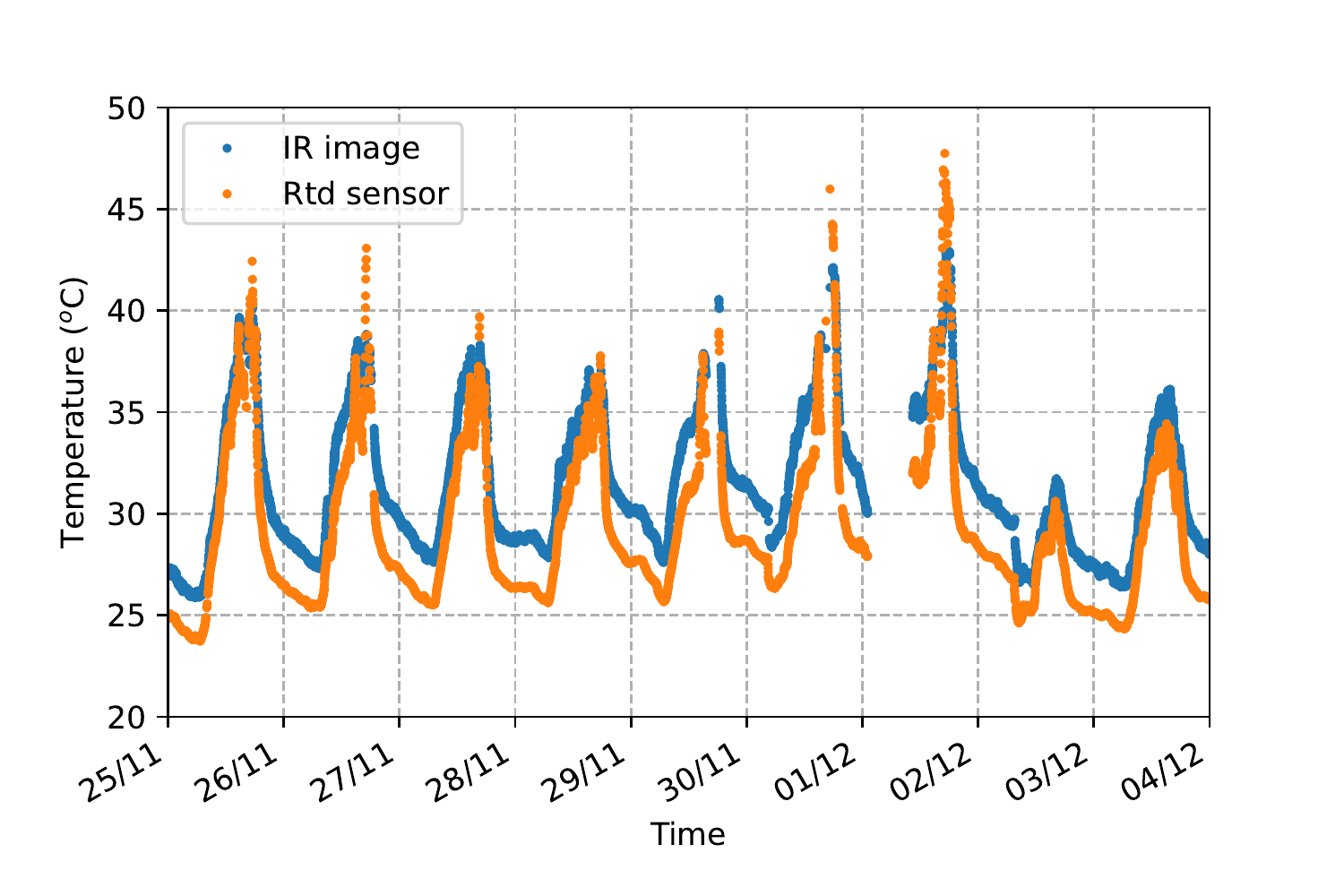}
         \caption{Surface temperature}
     \end{subfigure}
     \begin{subfigure}[b]{0.5\textwidth}
         \centering
         \includegraphics[width=\textwidth]{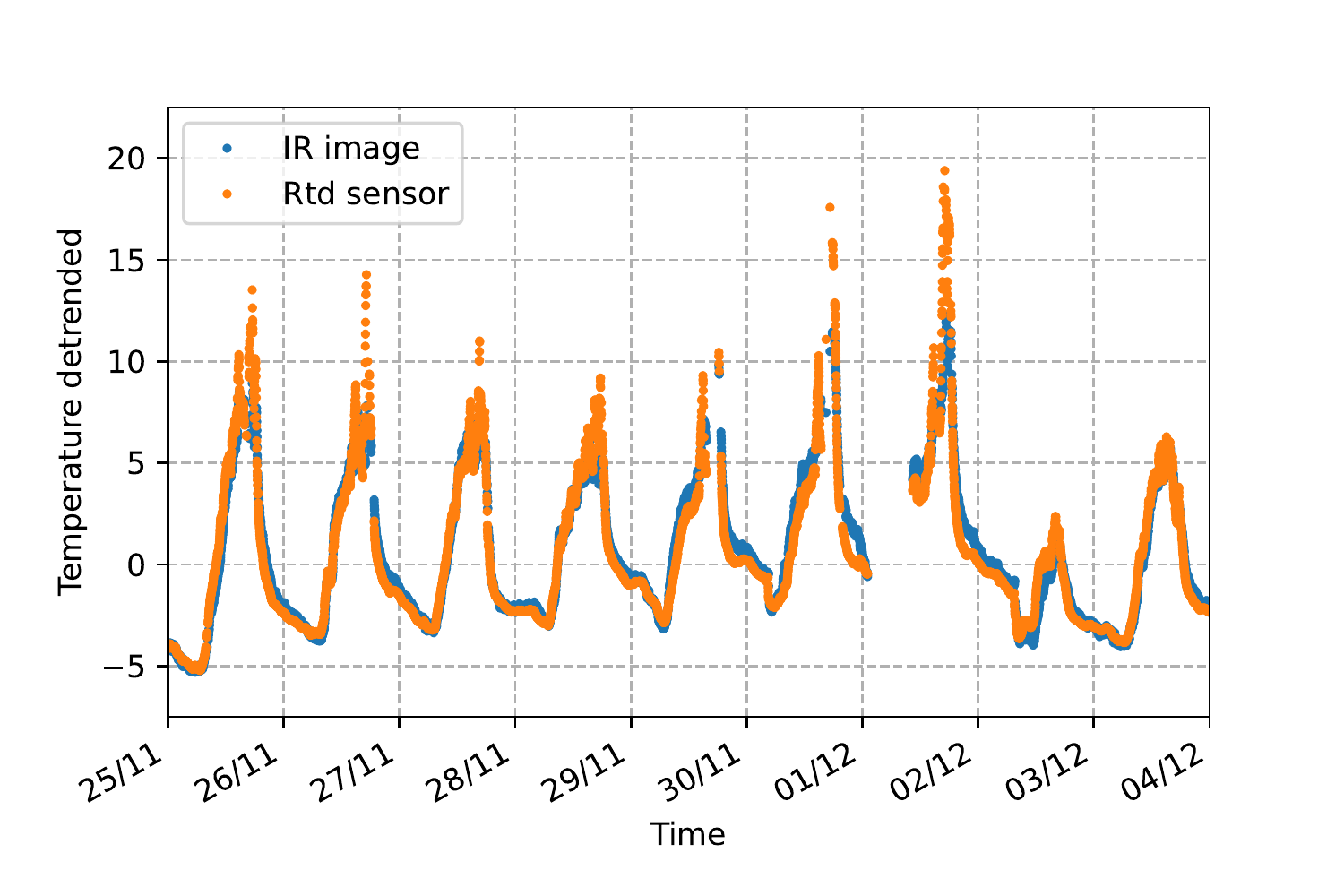}
         \caption{De-trended temperature}
     \end{subfigure}
        \caption{Plot showing the (a) surface temperature recorded using RTD sensor and thermal camera, and (b) corresponding de-trended temperature time-series.}
        \label{figure:ComparisonIRandRTD}
\end{figure}

The accuracy of the temperature data collected using the thermal image is compared against the surface temperature of the windows measured using the RTD sensors. Figure \ref{figure:ComparisonIRandRTD}(a) shows the temperature extracted from thermal images and the window surface temperature measured using the RTD sensor. It is observed that the temperature recorded using the IR camera is higher than the temperature measured using the RTD surface contact sensor. From Equation \ref{eq: eq2}, it is possible to infer that the camera output signal depends not only on the radiation from the object but also on the emissivity of the object, radiation from the atmosphere, and transmissivity of the atmosphere. Thus, the difference in the absolute value of the temperature could be because of the differences in the assumed emissivity value and the actual emissivity value of the glass window, radiation from the surrounding environment, and also changes in the ambient conditions. For better accuracy in the measured values, such as in the case of studies on urban heat island effect, using accurate values of emissivity of the objects, can help improve the measurement accuracy. Nevertheless, in this study, it is observed that the de-trended time series (Figure \ref{figure:ComparisonIRandRTD}(b)) of the surface temperature from the IR image and the RTD sensor are very similar. Here, the de-trending of the data is achieved by subtracting the least square fit to the data from the data itself. It is evident from the plot that even though the IR temperature may not measure the absolute surface temperature, the time-series pattern can be used to identify changes in the surface temperature and thereby the operational pattern.

\subsection{Air-conditioning usage}
In this section, the results from the analysis of centralized HVAC operation are presented first followed by analysis of the window AC unit operation. 

\subsubsection{Centralized HVAC operation}
\begin{figure}
     \centering
     \begin{subfigure}[b]{0.45\textwidth}
         \centering
         \includegraphics[width=\textwidth]{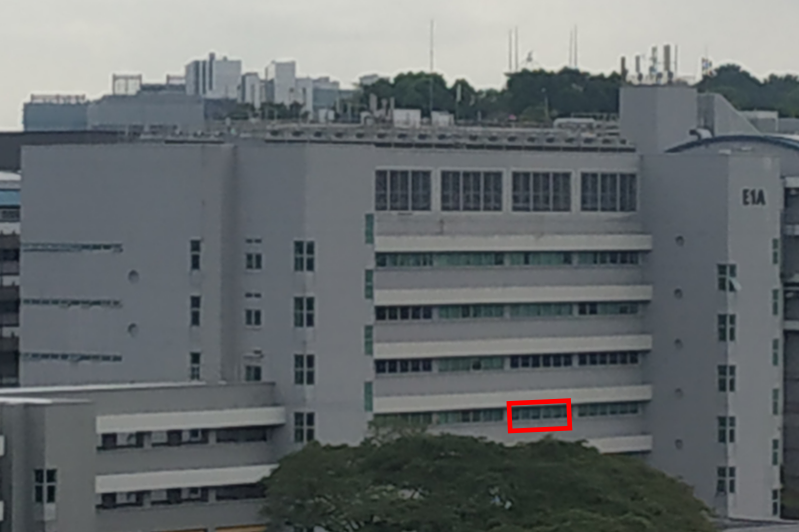}
         \caption{Location of region of interest (RoI)}
     \end{subfigure}
     \begin{subfigure}[b]{0.5\textwidth}
         \centering
         \includegraphics[width=\textwidth]{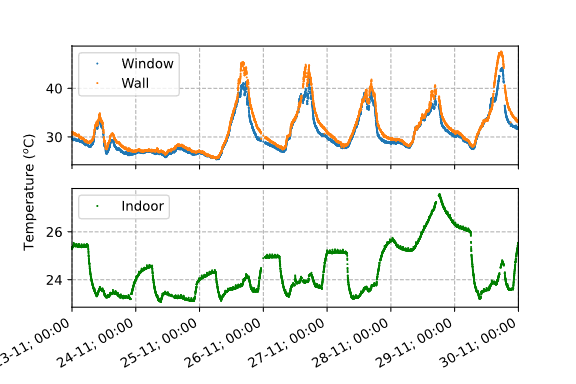}
         \caption{Window, wall and indoor temperature}
     \end{subfigure}
        \caption{(a) Digital image of the building under consideration, the red box indicates the window whose temperature is being analysed; (b) time-series plot of the temperature of wall and window extracted from IR image and the indoor dry-bulb air temperature measured using indoor sensor.}
        \label{figure:ubi_thermal}
\end{figure}

Figure \ref{figure:ubi_thermal}(a) shows the digital image of the building that is cooled through the centralised air conditioning system and the red box in the figure is the location of the office space where indoor sensor for monitoring dry-bulb temperature is installed. Figure \ref{figure:ubi_thermal}(b)(top) shows the surface temperature of window and wall measured using the thermal camera and Figure \ref{figure:ubi_thermal}(b)(bottom) shows the indoor dry-bulb temperature measured using the indoor monitoring sensor. As observed from the image, the temperature of the window is lower than the temperature of the wall and this difference is higher during the day. However, both the wall and window temperature time series appear to have a similar time series pattern. Whereas, the indoor dry-bulb temperature time-series pattern is quite different from the measured surface temperature of the window and wall. For instance, every day except for Saturday and Sunday, a sudden drop in the indoor temperature is observed at 6:00 in the morning and a gradual increase in temperature at 22:00 in the night. Similar behaviour is observed for Saturday, except the increase in temperature occurs at 18:00 in the evening instead of 22:00 in the night. From the indoor temperature profile, the instance corresponding to a sudden decrease in temperature corresponds to the time when the HVAC system is switched `on' and the instance corresponding to a gradual increase is the time HVAC system is switched `off'. 

The external surface heat balance is given as the sum of absorbed direct and diffuse solar radiation heat flux, net longwave thermal radiation flux exchange with air and surroundings, convective flux exchange with outside air and conduction heat flux into the wall. Thus the temperature of the window and wall surfaces exposed to the outdoor environment depends on several factors such as the outdoor temperature, the indoor temperature, temperature of the objects surrounding it, wind speed and direction, solar radiation, material properties of the surface, and the thickness of the surface. Any change in the external or indoor temperature or environmental factors would result in a change in the temperature of the window and the wall. For the wall and the window at the same location (exposed to similar environmental conditions), the rate of change in the temperature of the window and the wall would be different due to the differences in the thermal conductivity and the thickness of the material. Thus by observing the differences in the rate of change in the temperature of the wall and the window simultaneously, sudden changes in the indoor temperature can be detected. 

\begin{figure}
     \centering
     \begin{subfigure}[b]{0.5\textwidth}
         \centering
         \includegraphics[width=\textwidth]{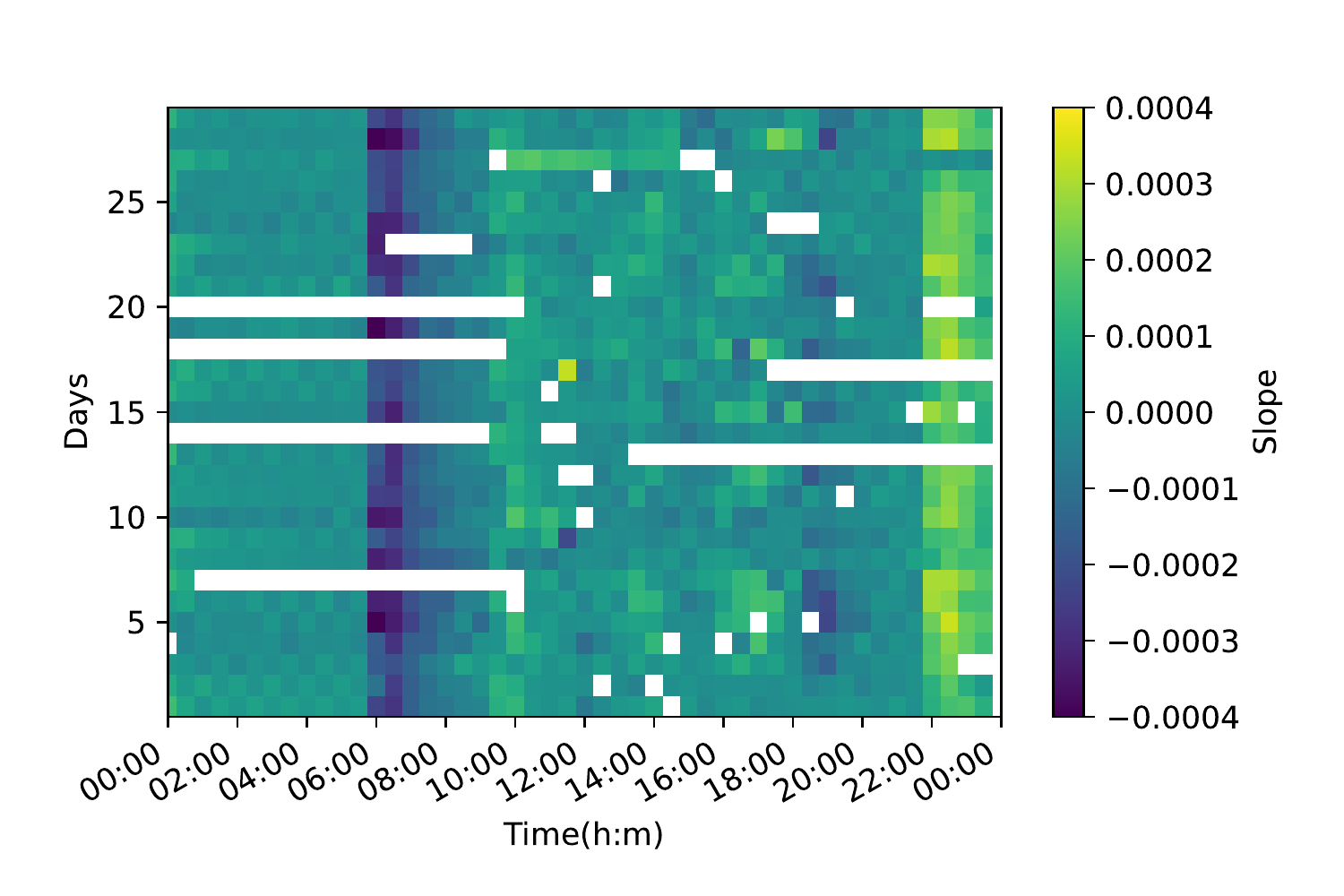}
         \caption{Dry-bulb indoor air temperature}
     \end{subfigure}
     \begin{subfigure}[b]{0.5\textwidth}
         \centering
         \includegraphics[width=\textwidth]{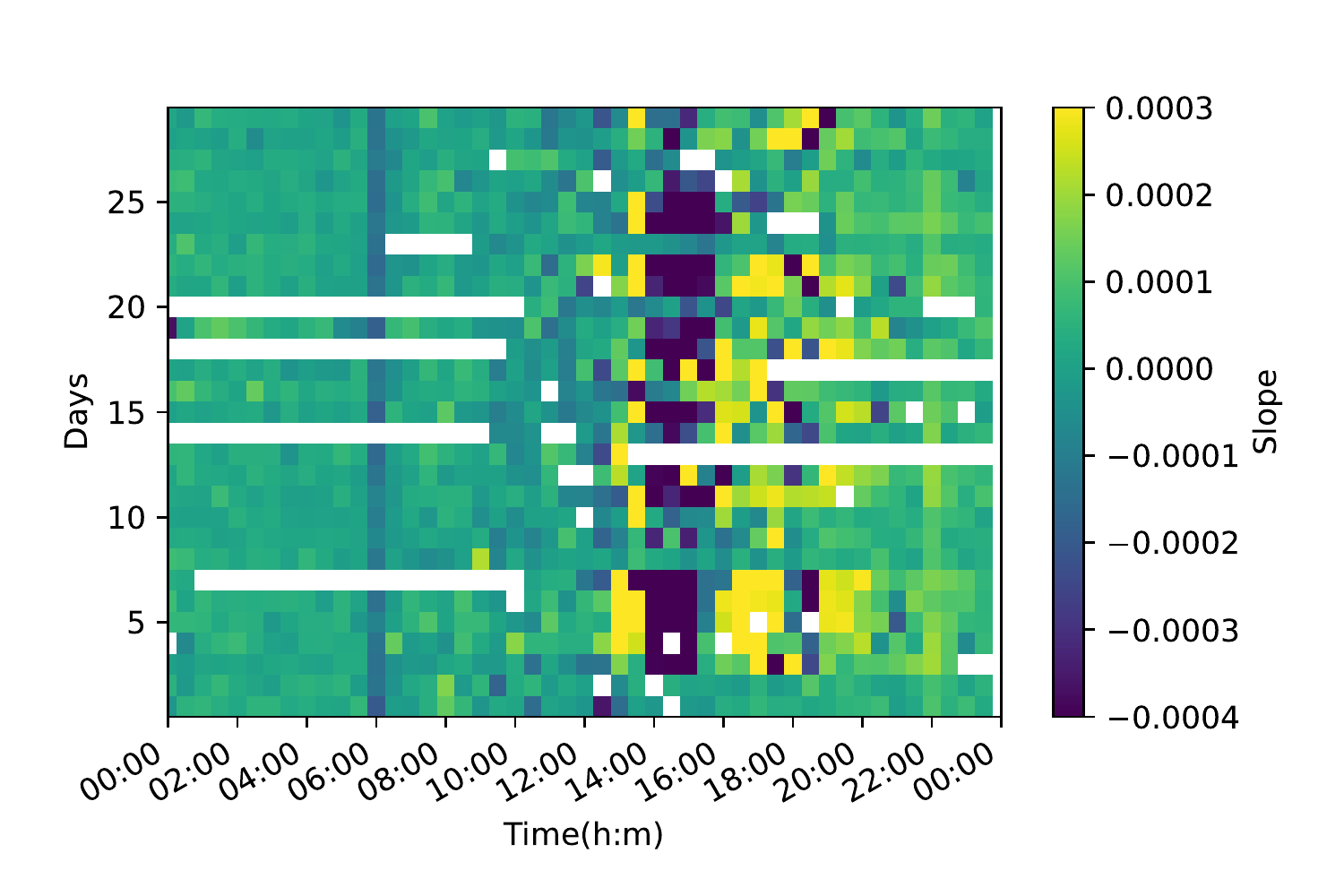}
         \caption{Thermal image}
     \end{subfigure}
        \caption{Heat-map showing the (a) the slope of detrended indoor dry-bulb air temperature for the weekdays, and (b) slope of the detrended temperature difference between the window and wall obtained from the IR image.}
        \label{figure:ubi_thermal_slope_weekday}
\end{figure}

Figure \ref{figure:ubi_thermal_slope_weekday}(a) and (b) shows the rate of change in the indoor dry bulb temperature measured using the temperature sensor and the rate of change in the difference between the window and the wall de-trended temperature from the IR image respectively. The rate of change in the de-trended temperature time series is estimated at every 30 minutes time interval. As shown in in the heatmap of Figure \ref{figure:ubi_thermal_slope_weekday}(a) at 6:00 in the morning, the rate of change in the temperature with time changes from a value close to 0 to a negative value. This is because the HVAC system is switched `on' and the indoor temperature decreases to the set-point temperature in a short time period. A similar change is observed in the heatmap shown in Figure \ref{figure:ubi_thermal_slope_weekday}(b) for the rate of change extracted from the IR image.

\begin{figure}
     \centering
     \begin{subfigure}[b]{0.55\textwidth}
         \centering
         \includegraphics[width=\textwidth]{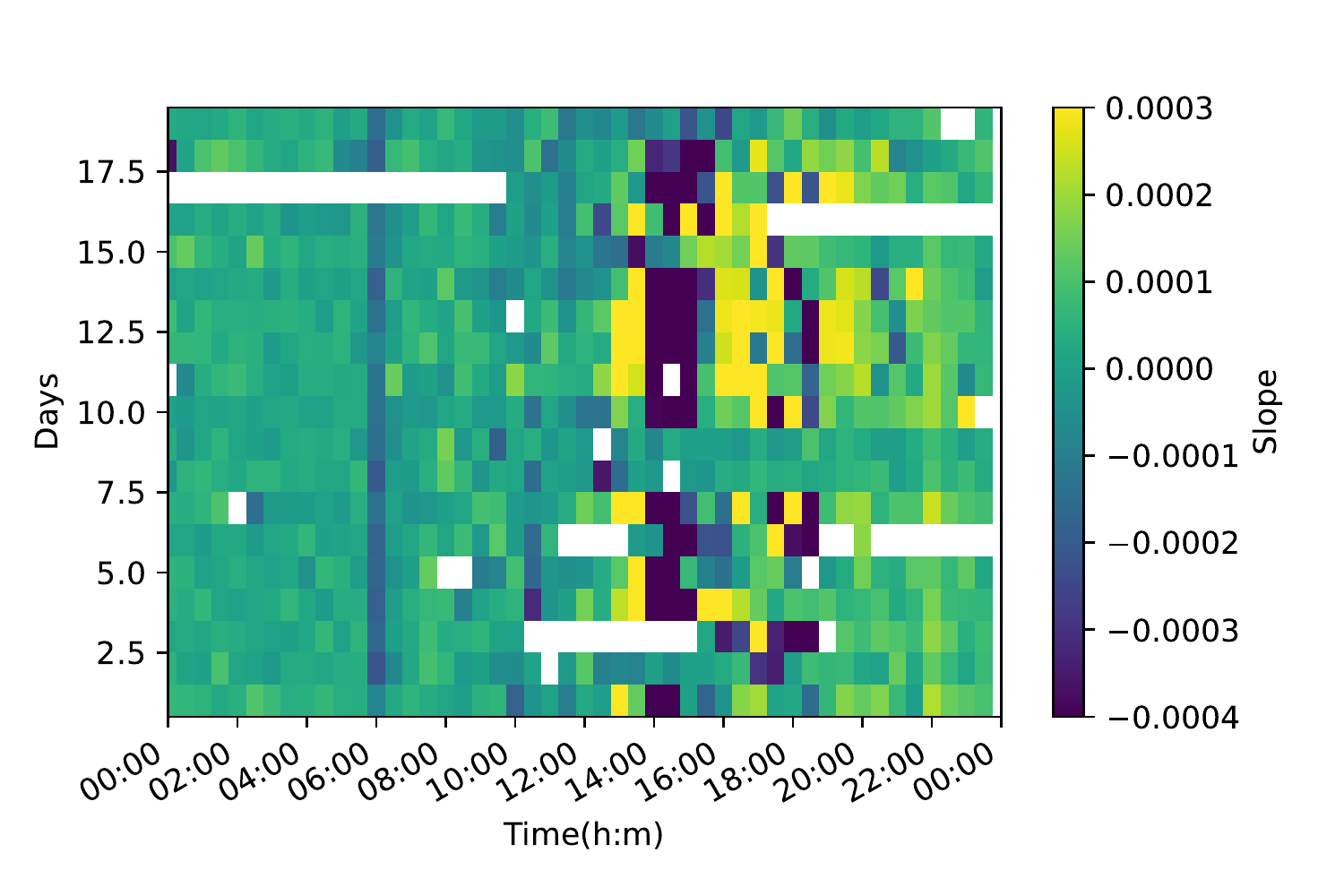}
         \caption{Slope from thermal image}
     \end{subfigure}
     \begin{subfigure}[b]{0.55\textwidth}
         \centering
         \includegraphics[width=\textwidth]{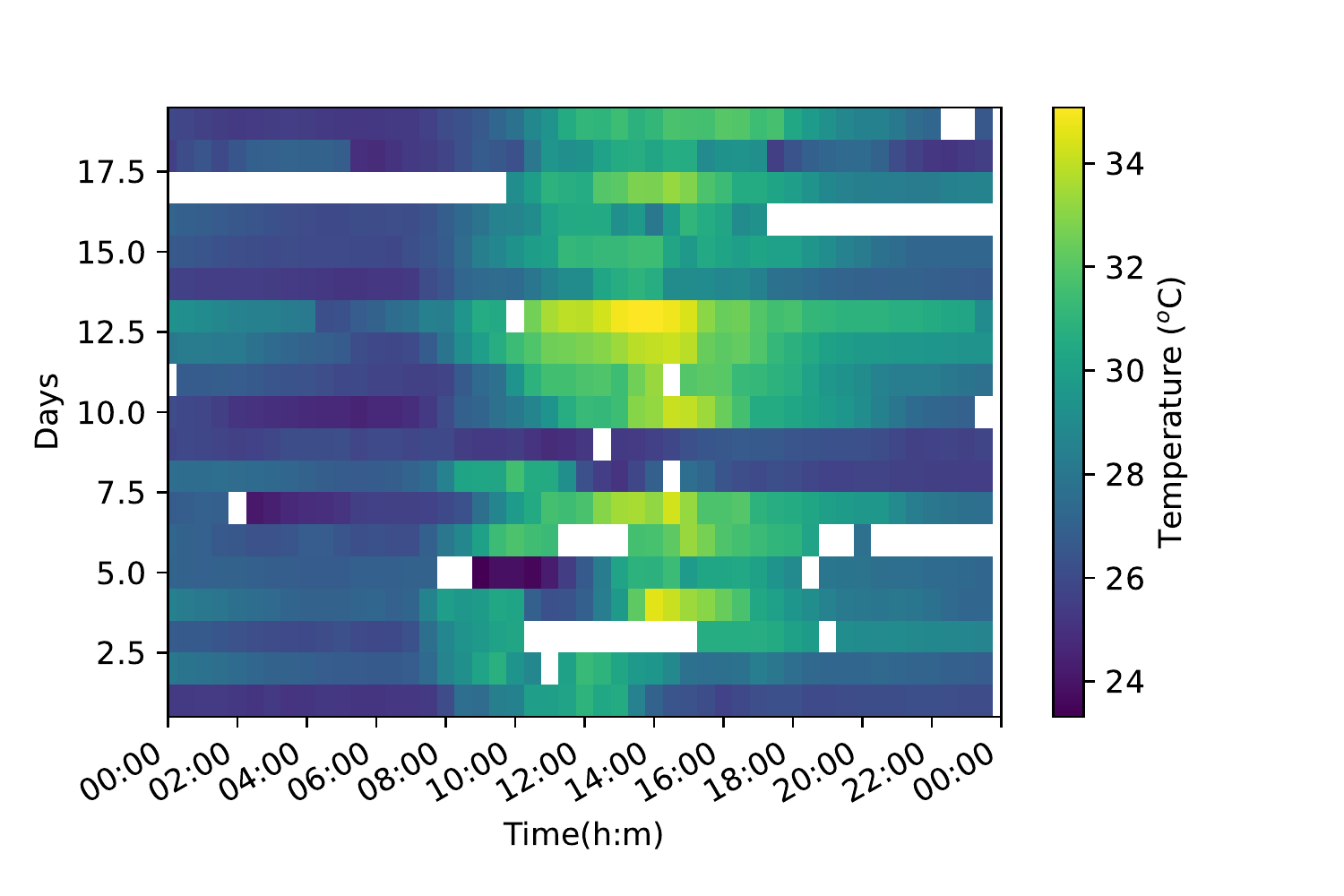}
         \caption{Outdoor temperature}
     \end{subfigure}
     \begin{subfigure}[b]{0.55\textwidth}
         \centering
         \includegraphics[width=\textwidth]{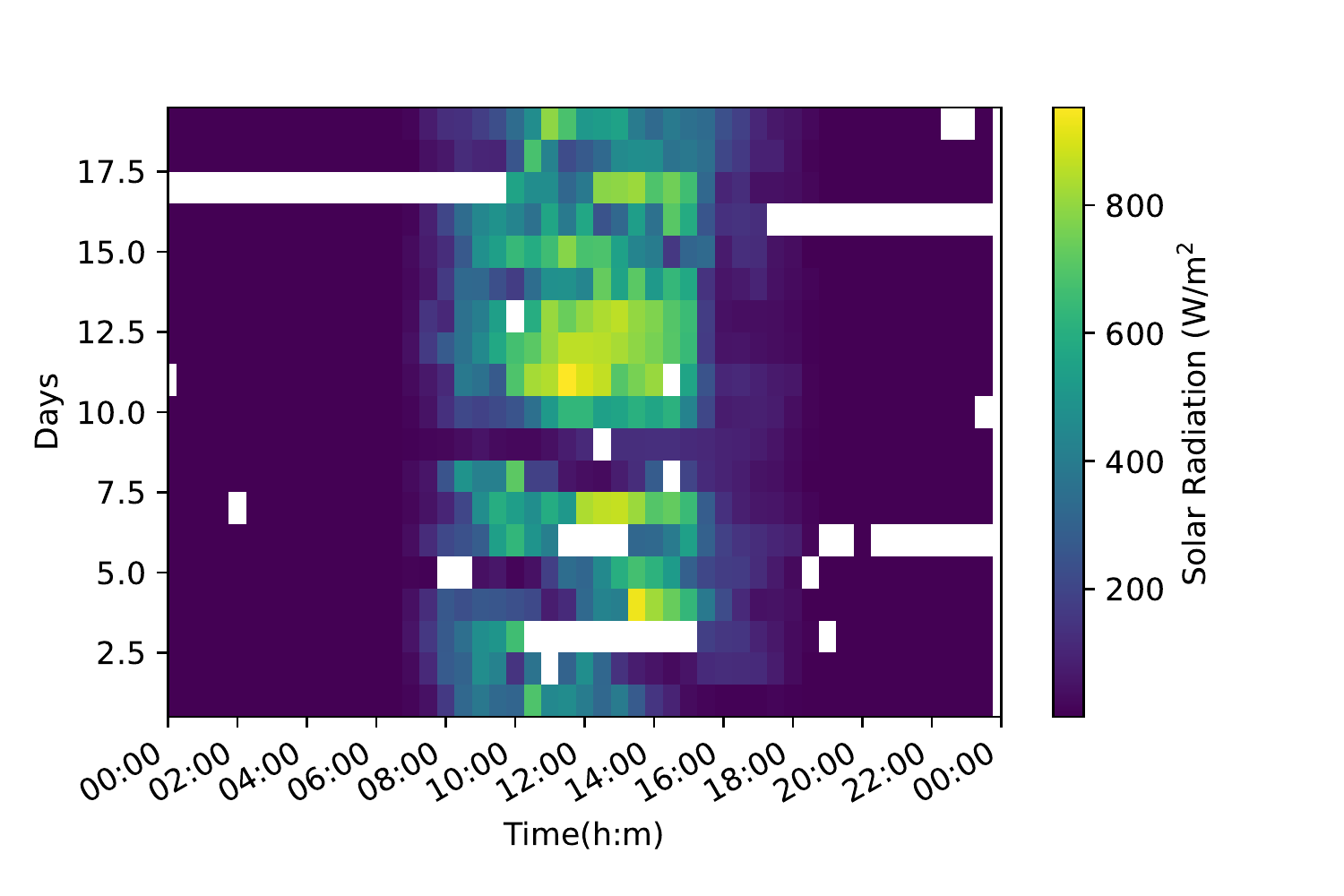}
         \caption{Solar radiation}
     \end{subfigure}
        \caption{Heat-maps showing (a) the slope of the detrended temperature difference between the window and wall obtained from the IR image for (b) outdoor temperature and (c) solar radiation measured at the weather station located at the observatory.}
        \label{figure:compare_with_outdoor_temperature}
\end{figure}

In the same manner, at 22:00, the rate of change in the temperature with time changes from a value close to zero to a positive value for both the heatmaps shown in Figure \ref{figure:ubi_thermal_slope_weekday}(a) and (b). This is the time the HVAC system is switched `off' and the temperature of the indoor space increases. Thus, it is possible to conclude that by observing the temperature of the windows and the walls over a span of few days, it is possible to obtain the operational behaviour of the centralised HVAC system. It is noted that unlike the heat map in Figure \ref{figure:ubi_thermal_slope_weekday}(a), which has a constant value during the day, the heat map in Figure \ref{figure:ubi_thermal_slope_weekday} shows a slope changes during the day. This is mainly due to the changes in outdoor temperature and solar radiation. This can be observed from Figure \ref{figure:compare_with_outdoor_temperature}, which depicts the heatmap for the rate of change in temperature difference between the window and wall, the outdoor temperature and solar radiation. On the days with a significant increase in the  outdoor temperature and solar radiation, significant changes in the slope are observed for the window-wall temperature slope during the day time. While for the days without significant changes in temperature and radiation (mainly cloudy and rainy days) the changes in the slope during the day is not significant. Thus, from the heatmap of the slope of the temperature difference between the wall and window it is possible to identify not just the operational pattern, but also rainy and sunny days. In the subsequent section, the operational behaviour of window AC units is discussed in detail.

\subsubsection{Window air-conditioning unit operation}
\begin{figure*}
     \centering
     \begin{subfigure}[b]{0.55\textwidth}
         \centering
         \includegraphics[width=\textwidth]{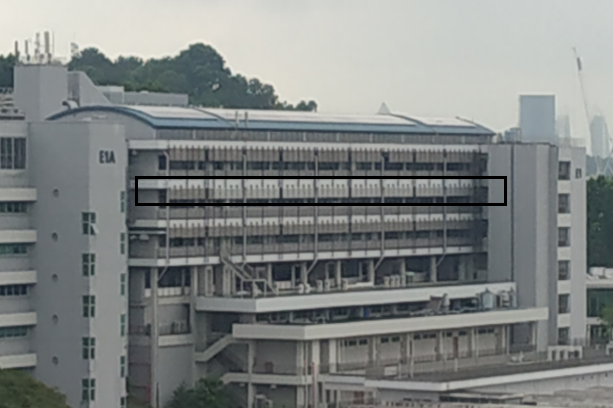}
         \caption{Digital image of the building}
     \end{subfigure}
     \begin{subfigure}[b]{0.55\textwidth}
         \centering
         \includegraphics[width=\textwidth]{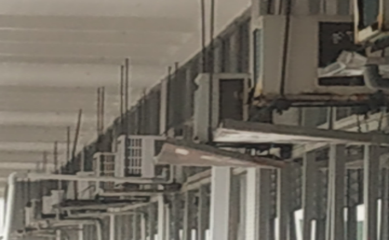}
         \caption{Digital image of the AC units}
     \end{subfigure}
     \begin{subfigure}[b]{0.55\textwidth}
         \centering
         \includegraphics[width=\textwidth]{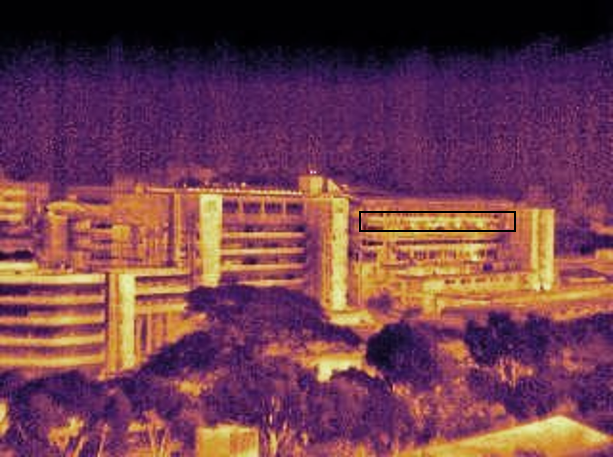}
         \caption{IR image (bright spots indicate the AC units}
     \end{subfigure}
        \caption{(a) Digital image showing the location of the window AC unit in building 3, (b) Digital image showing the window AC unit ; and (c) Bright spots in the IR image are the window AC units.}
        \label{figure:AC_unit_location}
\end{figure*}

\begin{figure}
     \centering
     \begin{subfigure}[b]{0.5\textwidth}
         \centering
         \includegraphics[width=\textwidth]{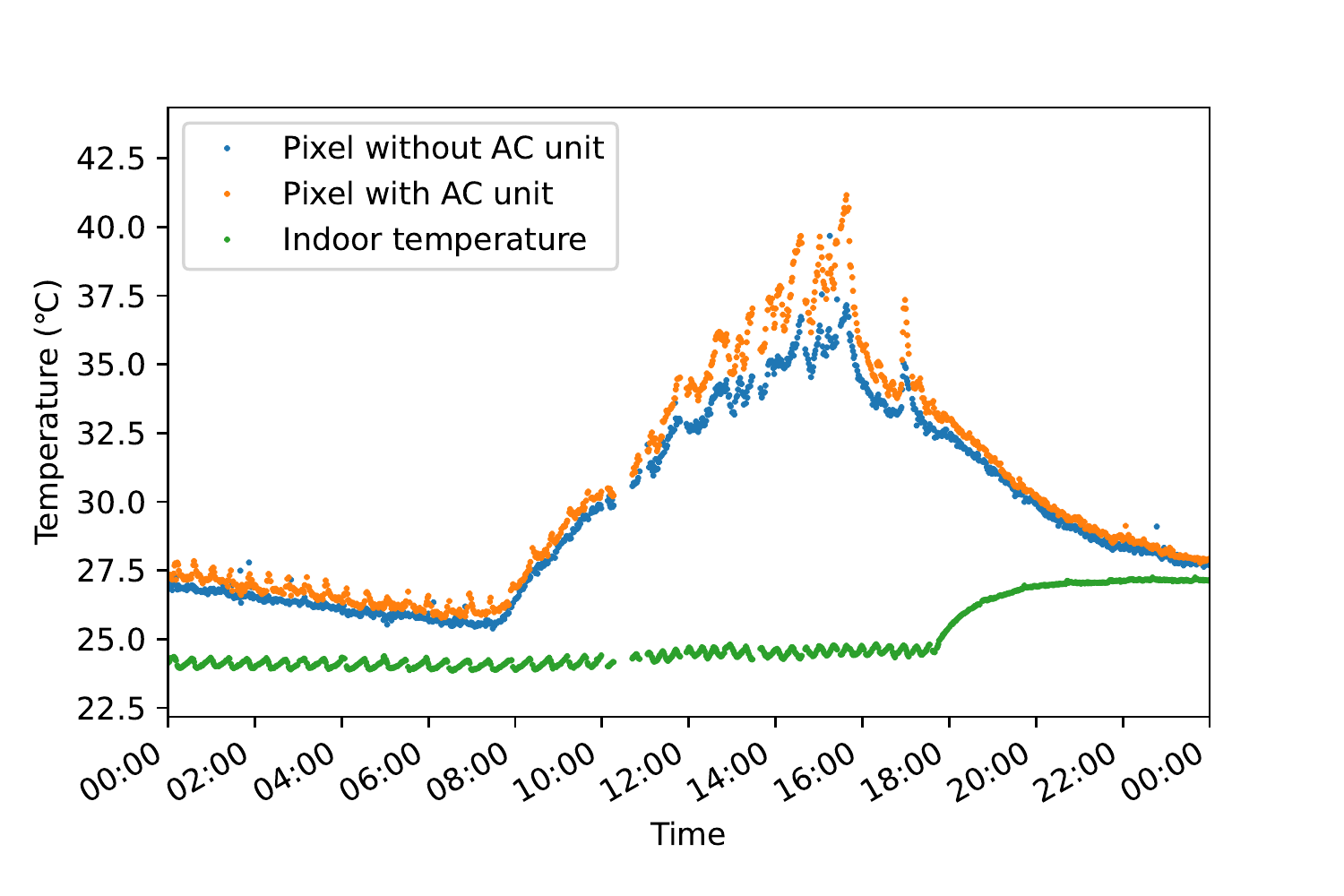}
         \caption{Temperature from thermal image and indoor sensor}
     \end{subfigure}
     \begin{subfigure}[b]{0.5\textwidth}
         \centering
         \includegraphics[width=\textwidth]{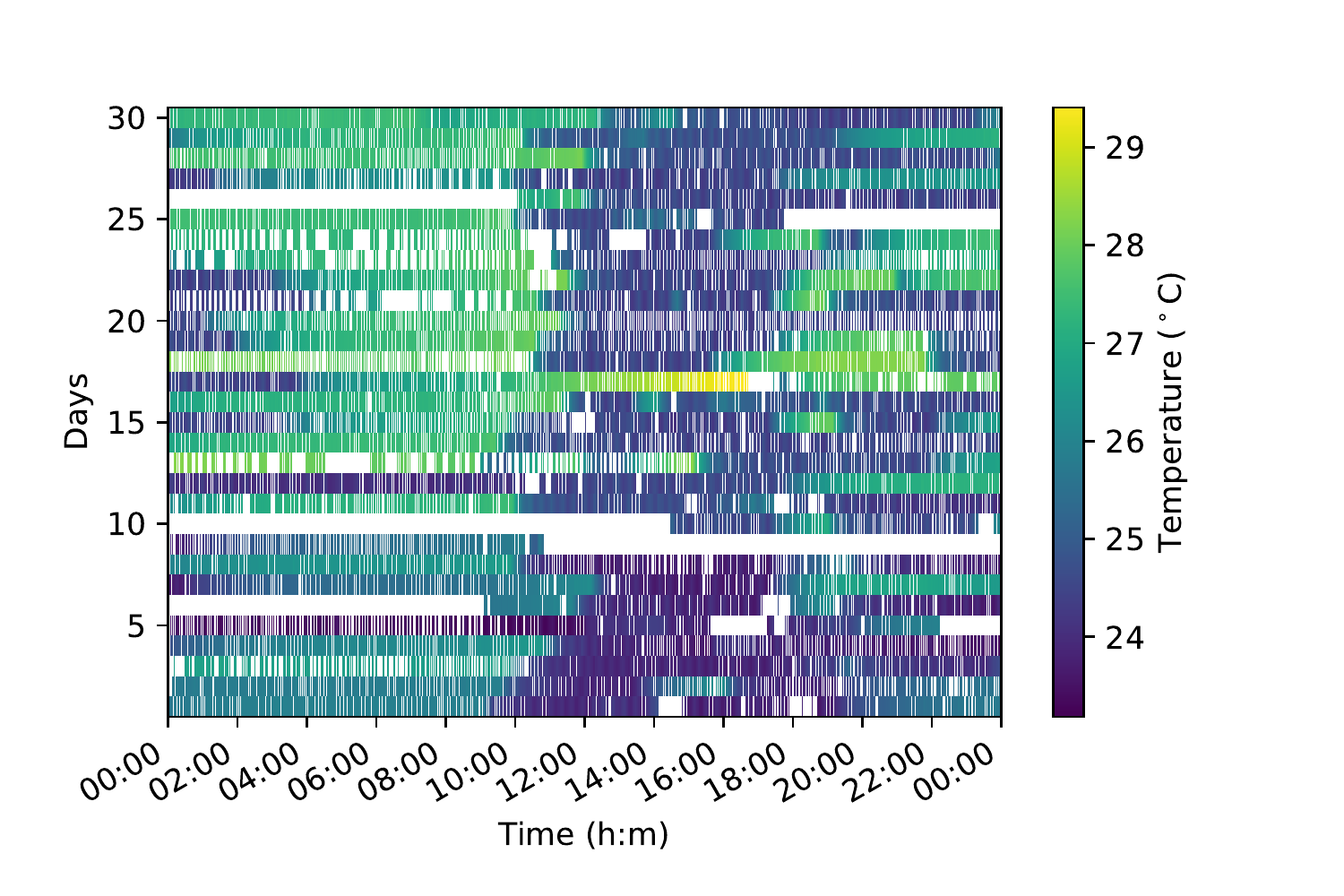}
         \caption{Indoor dry-bulb air temperature}
     \end{subfigure}
     \begin{subfigure}[b]{0.5\textwidth}
         \centering
         \includegraphics[width=\textwidth]{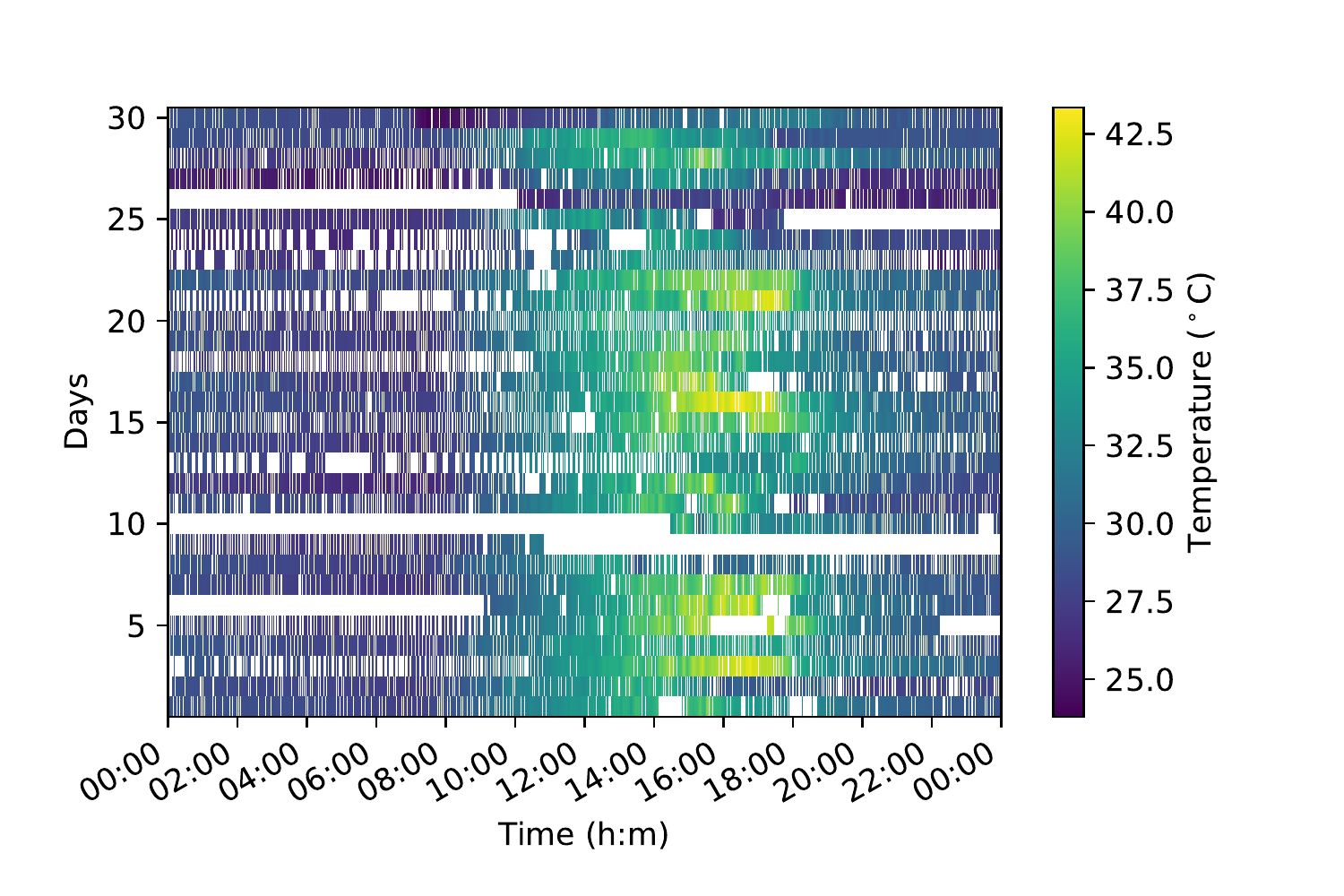}
         \caption{AC unit temperature - IR image}
     \end{subfigure}
        \caption{(a) Plot showing the time-series recorded for one day using thermal image and indoor temperature sensor, Heatmaps showing the temperature measured over a duration of few days of the (b) indoor office space measured using indoor temperature sensor and (c) pixel corresponding to AC condenser unit from thermal image.}
        \label{figure:AC_unit_operation}
\end{figure}

Figure \ref{figure:AC_unit_location}(a) and (b) shows the digital image of the location of the  window AC units (black box) and series of window AC units at the location respectively. Figure \ref{figure:AC_unit_location}(c) shows the corresponding IR image of the building. A bright spot is observed in the IR image at the location of the window AC units (see black box region in the image). The bright spot indicates that the AC unit is in operation, while a non-bright region along the same row indicates either the AC unit is not in operation or a pixel without any AC unit at that location. To better understand the variation in the temperature pattern during the AC unit operation, the temperature time series of the AC condenser unit pixel from the IR image is extracted. Figure \ref{figure:AC_unit_operation}(a) shows the temperature pattern of the pixel unit with an AC unit and a non-AC unit along the same row highlighted using black box in \ref{figure:AC_unit_location}(a). It is observed that the temperature of the pixel corresponding to the AC unit is relatively greater than the temperature of the non-AC unit pixel. The bright spot in the IR image is a result of the higher temperature of the condenser unit during its operation. Though this could be an indication of the AC operation, however, during the day as the temperature of the surrounding environment increases to as high as 40$^{\circ}$C, it can become difficult to identify its operation just by observing the differences in the temperature. From the graph, it is also observed that, in addition to the temperature of the condenser unit being higher than the surrounding pixel, the temperature time series has a wave-like pattern. This wave-like pattern is mainly a result of the duty cycle of the AC condenser unit during its operation. 

In addition to the temperature extracted from the IR images, Figure \ref{figure:AC_unit_operation}(a) compares the indoor dry bulb temperature (of the office space cooled by the same AC unit) with the AC condenser unit temperature extracted from thermal image. From the graph, the time at which the AC unit is switched `off' is clearly identified as increase in the indoor dry bulb temperature. A similar change or decrease in the indoor temperature is observed when the AC unit is switched `on' from the indoor temperature profile. However, a similar change (sudden decrease or increase) in the temperature profile of the condenser unit extracted from the IR image may or may not be clearly observed. Figure \ref{figure:AC_unit_operation}(b) and (c) shows the heat map of the indoor temperature of the office space and the temperature of the pixel with the AC unit extracted from the IR image respectively for a longer duration of time. A fringe-like pattern is observed in both the temperature heat maps, which is indicative of the AC unit operation. This shows that the operational pattern of AC units can be detected by identifying the duty cycle from the temperature time series. 

\begin{figure*}
     \centering
     \begin{subfigure}[b]{0.7\textwidth}
         \centering
         \includegraphics[width=\textwidth]{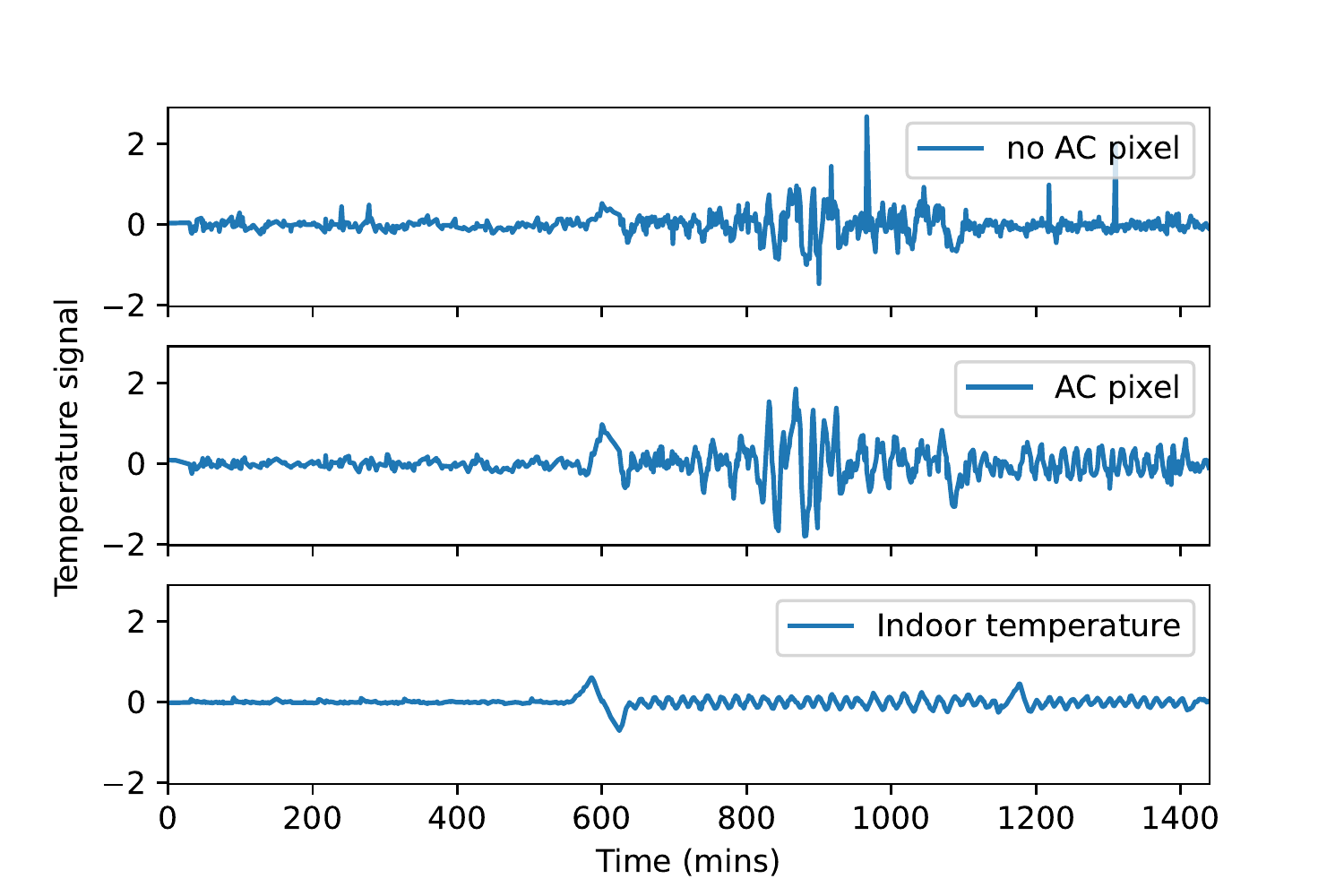}
         \caption{Moving averaged temperature}
     \end{subfigure}
     \begin{subfigure}[b]{0.7\textwidth}
         \centering
         \includegraphics[width=\textwidth]{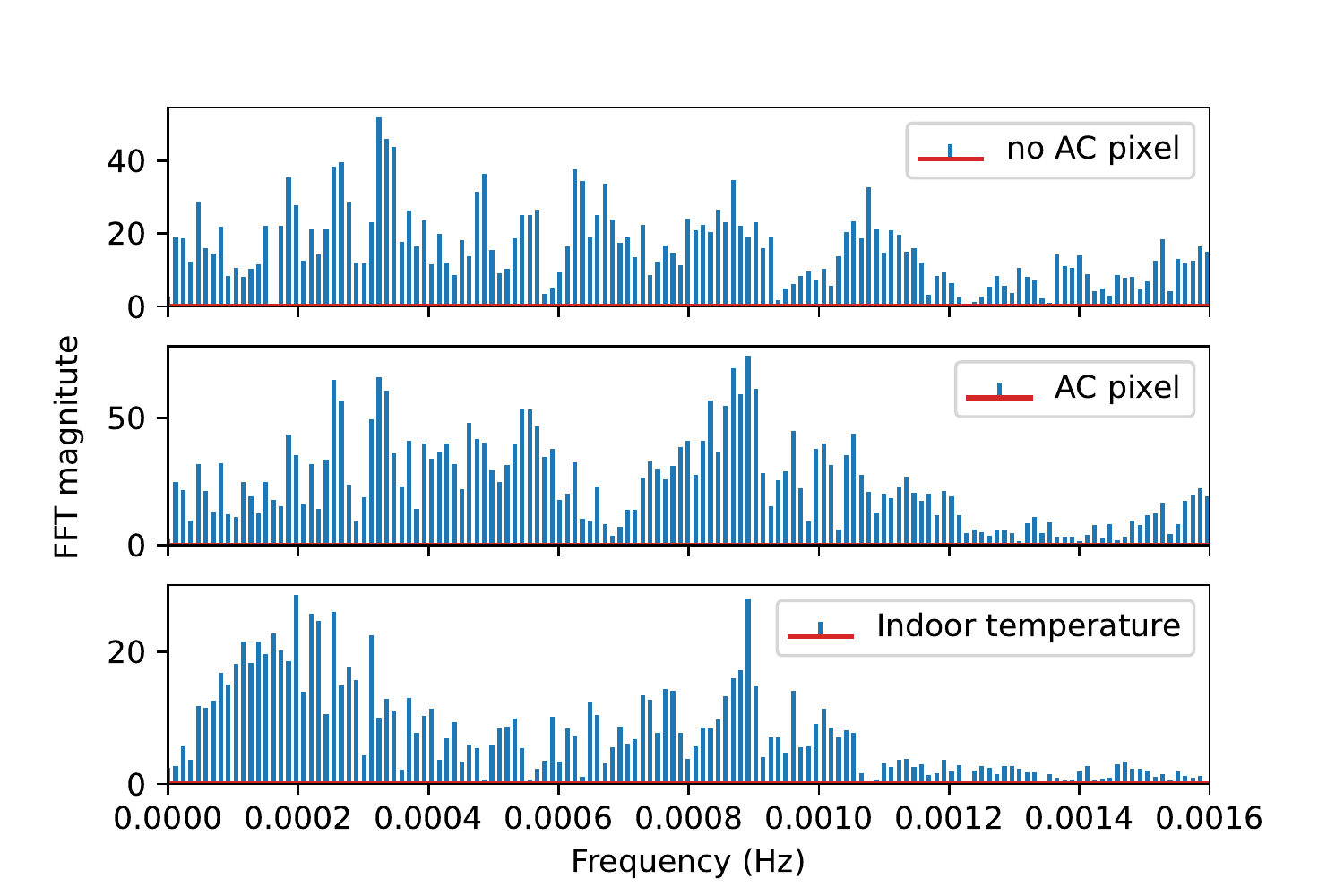}
         \caption{Fast fourier transform (FFT)}
     \end{subfigure}
        \caption{Plots showing (a) the moving averaged temperature signal and (b) its corresponding fast fourier transform.}
        \label{figure:FFT}
\end{figure*}

A fast Fourier transform (FFT) is commonly used for identifying the frequency content of the time series with a wave-like pattern. Figure \ref{figure:FFT} shows the FFT of the moving averaged temperature time series of the AC unit, the non AC unit pixel from the IR image, and the indoor dry bulb temperature. It is observed from the FFT spectrum of the temperature time series of the pixel with and without AC unit, there exists a significant difference in the FFT magnitude for the frequency in the range of 0.0005 to 0.0011 (Hz). Also, comparing the FFT plots for the indoor temperature and the temperature time series corresponding to AC unit pixel, it is observed that there exists a dominant frequency in the same range, which is also the missing frequency in the case of the temperature of the pixel without AC unit. This indicates that the duty cycle of the AC unit has a frequency in the range of 0.0005 to 0.0011 Hz. Thus identifying the time of occurrence of this frequency in the time series can be indicative of the AC unit operation. However, the time operation during the day cannot be obtained directly from the FFT spectrum alone. A wavelet transform, on the other hand, is a handy mathematical tool that allows the identification of the time when a frequency occurs in a signal.

\begin{figure*}
     \centering
     \begin{subfigure}[b]{0.45\textwidth}
         \centering
         \includegraphics[width=\textwidth]{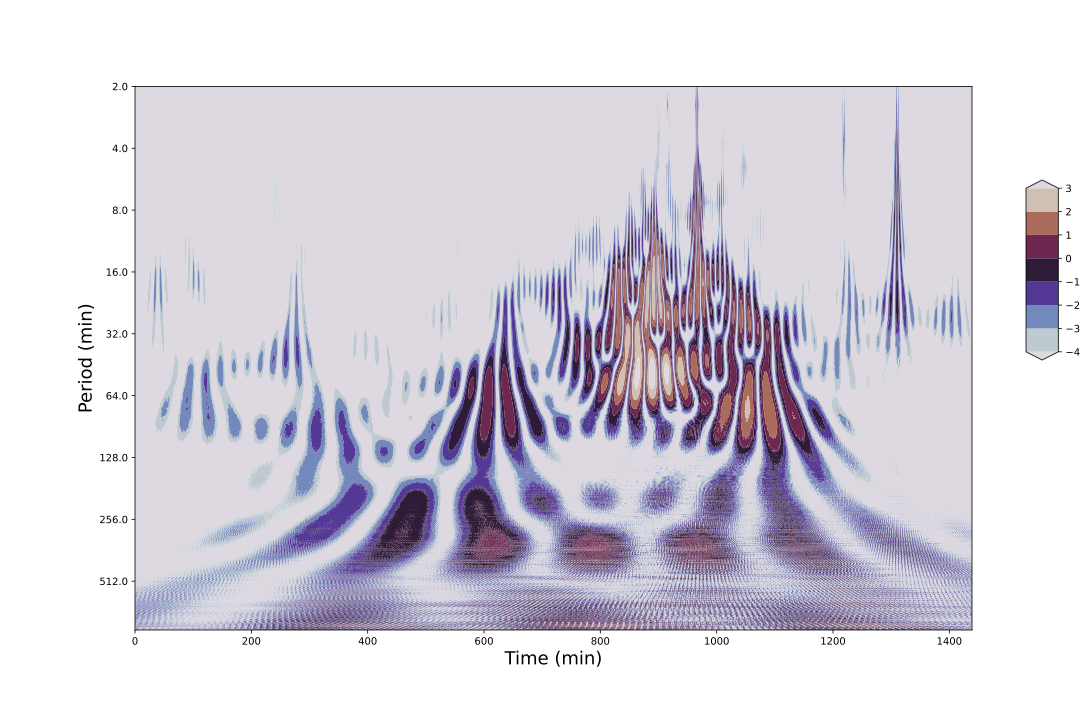}
         \caption{No AC unit thermal WT}
     \end{subfigure}
     \begin{subfigure}[b]{0.45\textwidth}
         \centering
         \includegraphics[width=\textwidth]{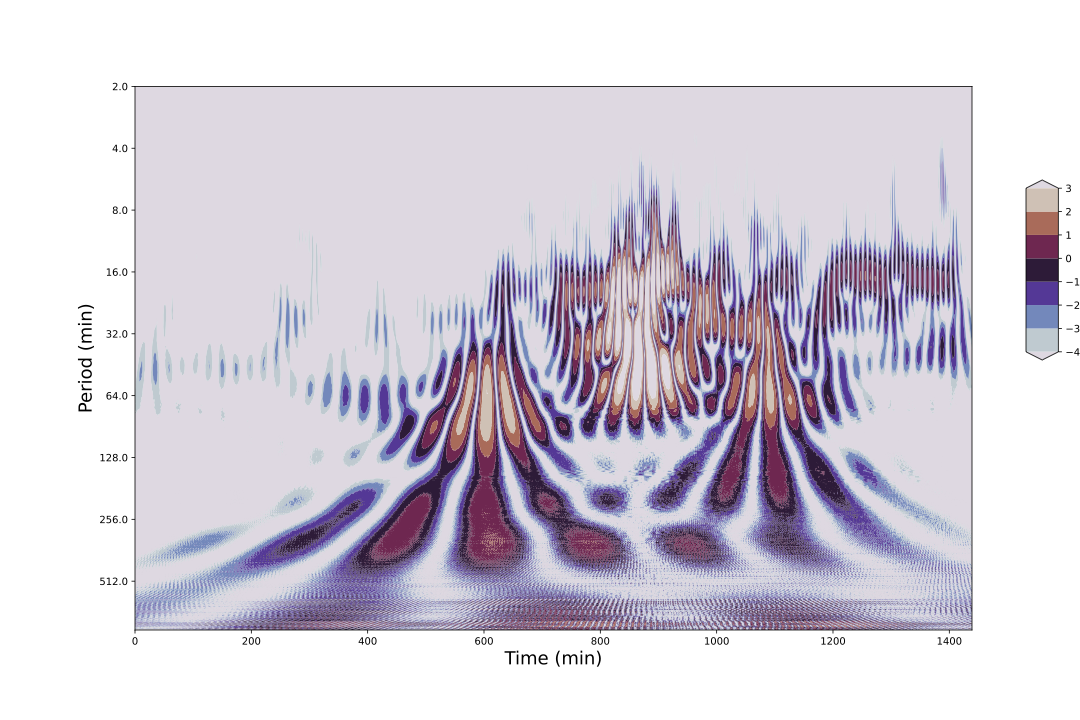}
         \caption{AC unit thermal WT}
     \end{subfigure}
     \begin{subfigure}[b]{0.45\textwidth}
         \centering
         \includegraphics[width=\textwidth]{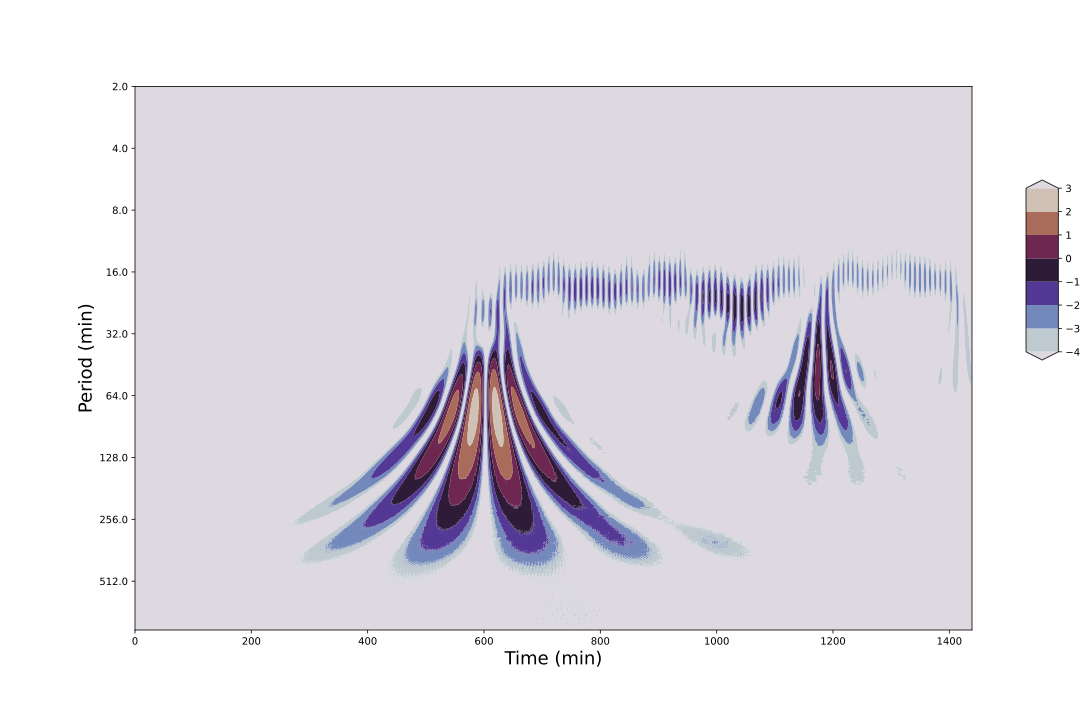}
         \caption{Indoor dry bulb temperature WT}
     \end{subfigure}
     \begin{subfigure}[b]{0.45\textwidth}
         \centering
         \includegraphics[width=\textwidth]{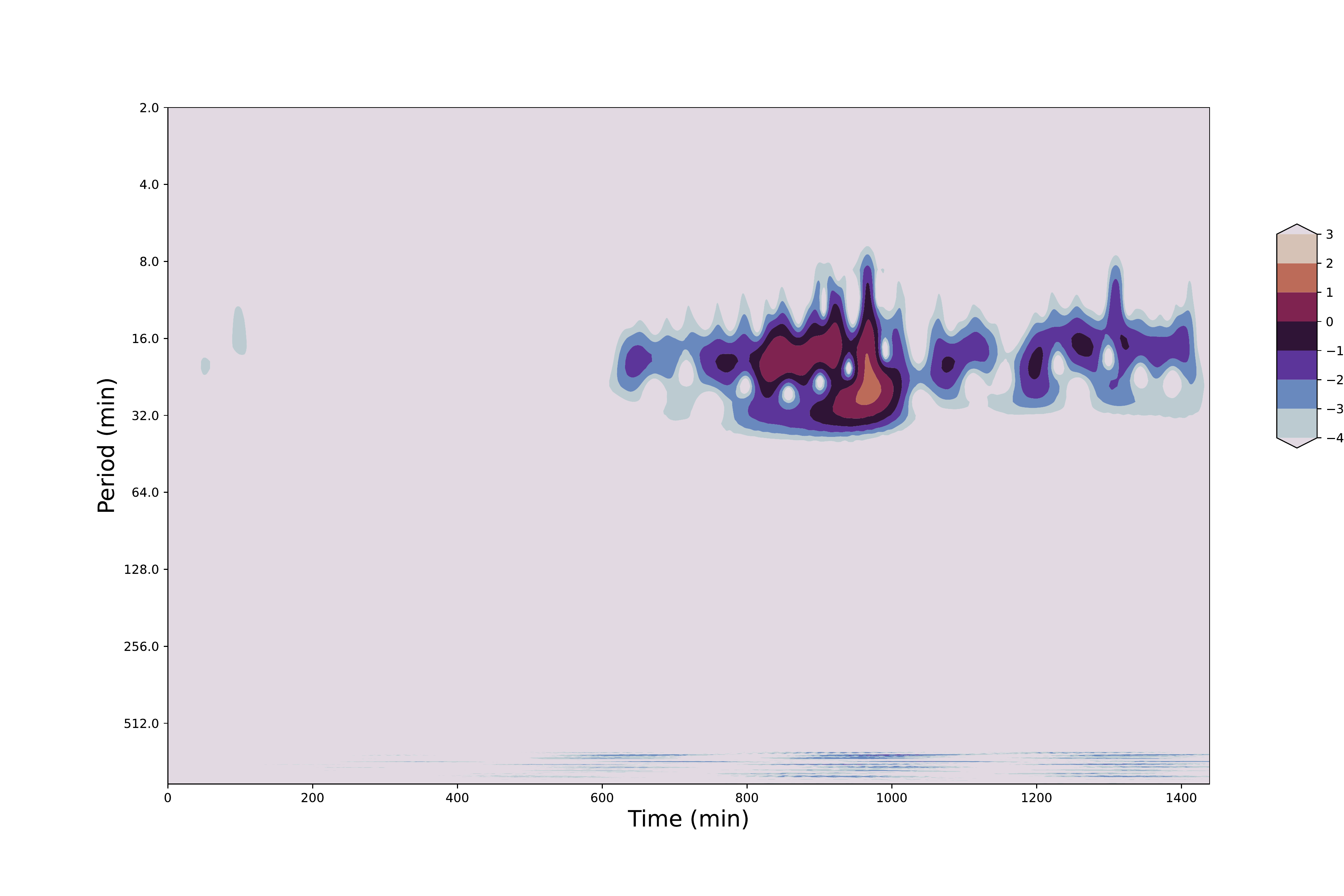}
         \caption{Cleaned AC unit thermal WT}
     \end{subfigure}     
        \caption{Wavelet transform spectrum of (a)temperature signal of pixel without AC condenser unit, (b)temperature signal of pixel with AC condenser unit, (c)dry-bulb indoor air temperature and (d)cleaned temperature signal of AC condenser}
        \label{figure:WT}
\end{figure*}

Figure \ref{figure:WT}(a) and (b) shows the wavelet transform spectrum of the time series corresponding to pixel unit without AC unit and with AC unit for the temperature signal shown in Figure \ref{figure:FFT}(a). It is observed that between the time stamp of 600 min to 1400 min the AC unit signal has a dominant frequency in the range of 16 min and 32 min, which indicates the AC unit is in operation. The frequencies other than the AC unit operation frequency could be due to the effect of external temperature changes, solar radiation, and changes in environmental factors. The AC condenser unit temperature signal is further cleaned to remove the frequencies other than the operating AC unit frequency, which can then be used to estimate the duration of AC usage from the wavelet spectrum. Figure \ref{figure:WT}(c) and (d) shows the wavelet transform spectrum of the indoor temperature and the frequency cleaned AC unit temperature signal. It can be observed that there exists a dominant operational frequency during the same time duration in the wavelet spectrum of the indoor temperature signal. Thus, the two spectrum can be used for comparing the accuracy in the prediction of the duration of operation of the AC unit. 

\begin{figure}
     \centering
     \begin{subfigure}[b]{0.5\textwidth}
         \centering
         \includegraphics[width=\textwidth]{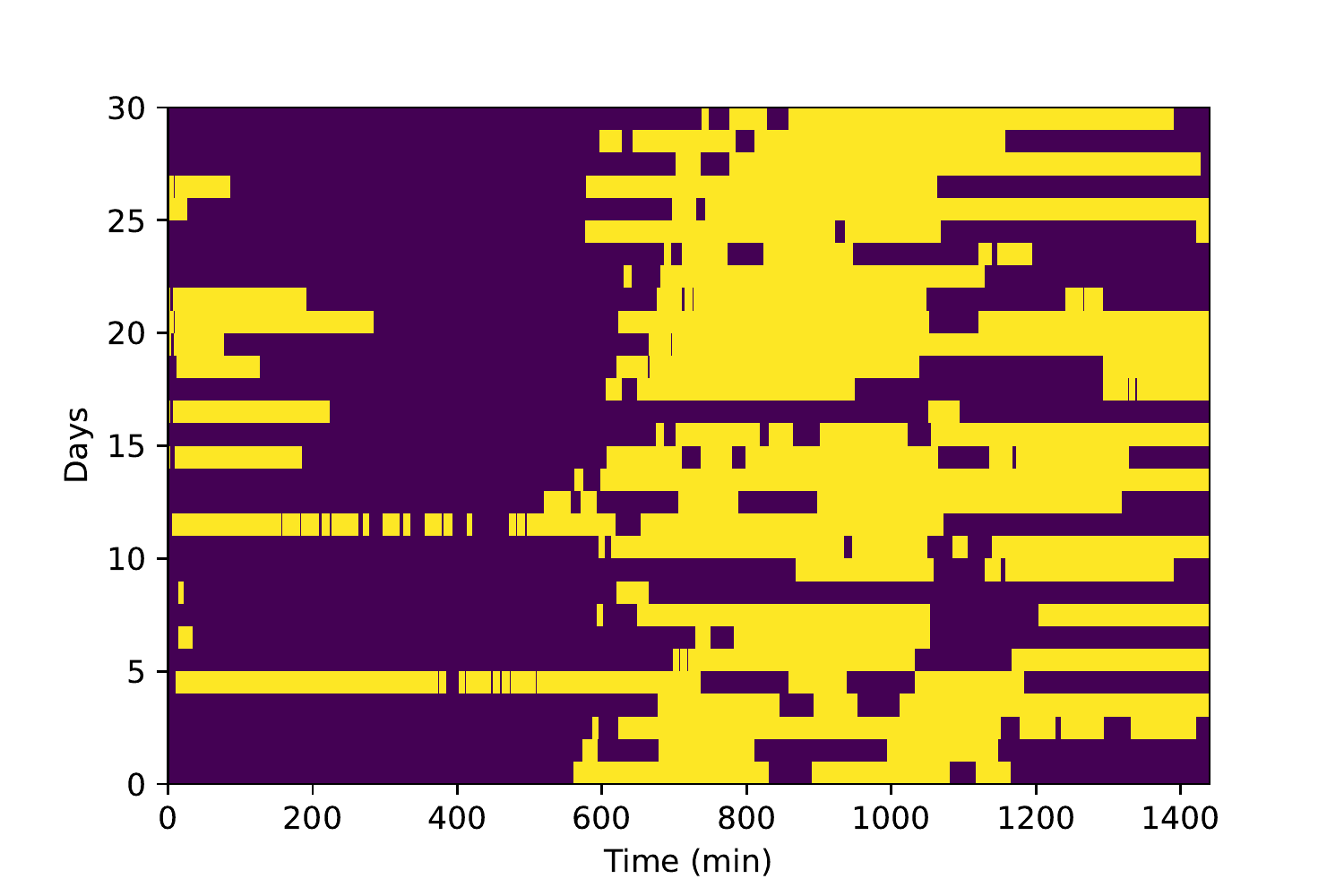}
         \caption{AC usage from indoor dry bulb temperature}
     \end{subfigure}
     \begin{subfigure}[b]{0.5\textwidth}
         \centering
         \includegraphics[width=\textwidth]{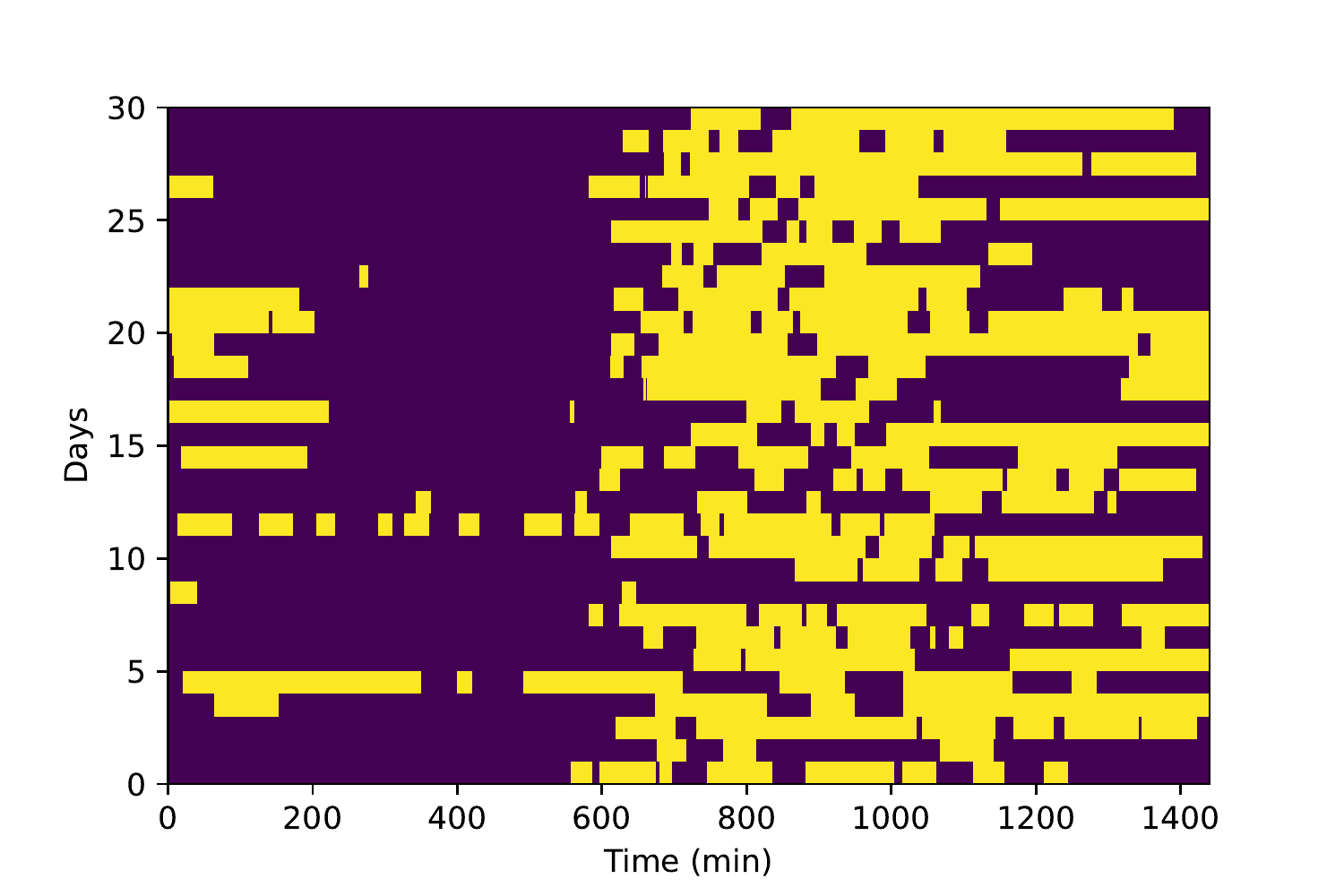}
         \caption{AC usage from IR temperature}
     \end{subfigure}
     \begin{subfigure}[b]{0.5\textwidth}
         \centering
         \includegraphics[width=\textwidth]{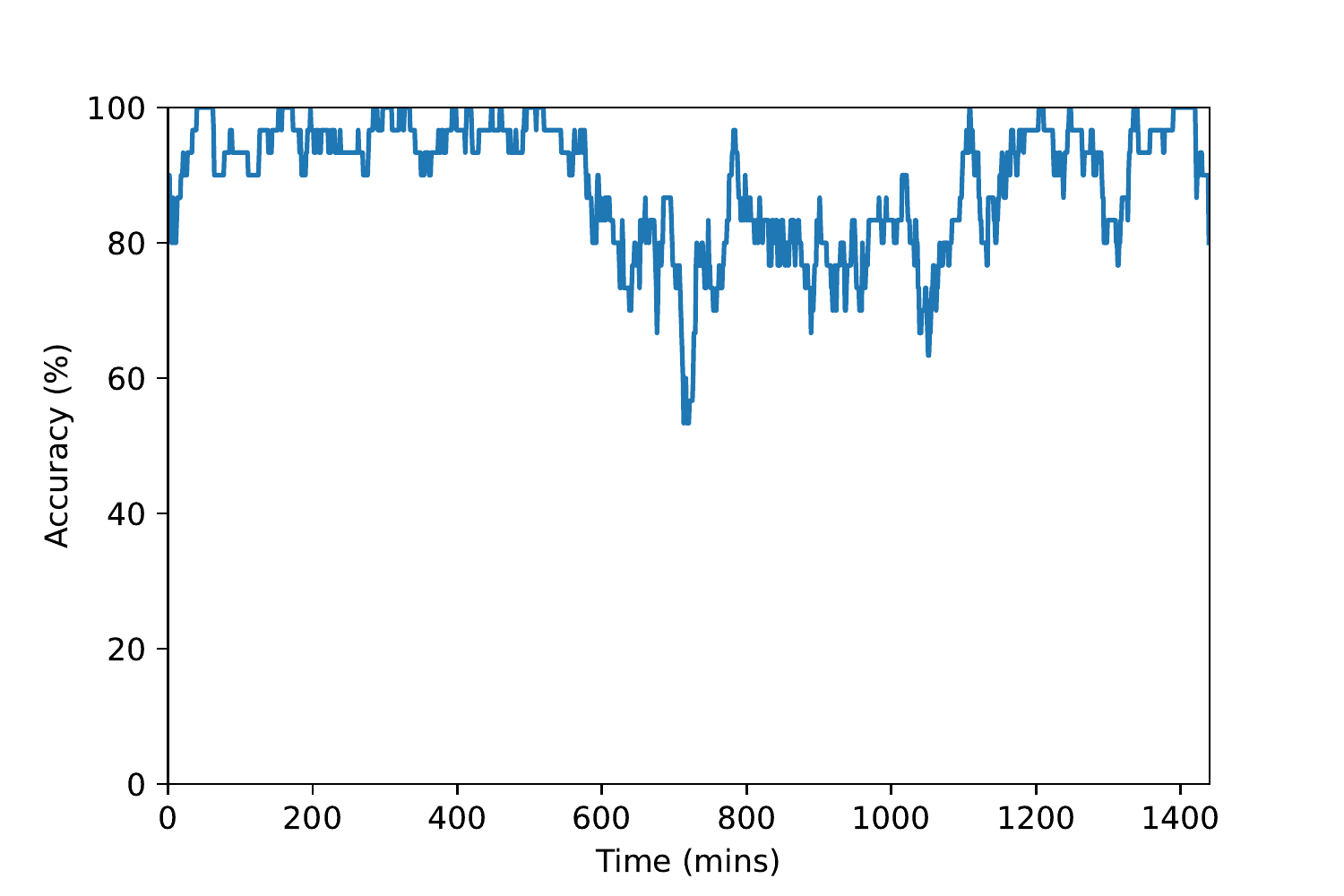}
         \caption{Prediction accuracy}
     \end{subfigure}
        \caption{Plot showing the duration of AC usage estimated from the wavelet spectrum of (a) dry-bulb indoor air temperature sensor signal and (b) thermal image respectively. The yellow regions represent the time when the AC was on during the day. (c) Plot showing the accuracy in prediction of AC usage during the day.}
        \label{figure:accuracy}
\end{figure}

Figure \ref{figure:accuracy}(a) and (b) shows the heatmaps of the duration of the AC unit usage extracted from the wavelet transform spectrum of the indoor temperature sensor signal and AC unit temperature from the thermal image respectively. The plots demonstrate that from the wavelet spectrum of the temperature signal of the condenser unit, it is possible to gain insights into the duration of operation of the AC units. As mentioned previously, the IR temperature is affected not just by the object under consideration but as well as external temperature, solar radiation, and emissivity of the objects. To study the effect of ambient temperature and solar radiation variation during the day, accuracy is predicted with time and shown in Figure \ref{figure:accuracy}(c). It is observed that prediction accuracy is high during the night and early morning between 8 pm to 10 am and decreases during the day. This is mainly due to the effect of very high solar radiation and outdoor temperature. Based on the accuracy results, it is recommended that for studies involving characterization of the AC units, signal extracted during the night time should be used.

\subsection{AC operational state detection and usage characterization}

Based on the Wavelet analysis it is realised that for characterization studies on AC unit it is best to consider the operation of the unit in the nighttime. To demonstrate this, similar to the residential buildings study in ~\cite{arjunan2021operational}, the operational state (`On' or `Off') of the AC units at night time is extracted from the individual temperature time series. A supervised method based on univariate $k$-means clustering is applied. The univariate $k$-Means algorithm partition the given one-dimensional data into $k$ clusters. 
In this study, $k$ is set with two to select the optimal decision boundary between the `On' or `Off' states in the given AC temperature time series data. 
As shown in Figure~\ref{figure:AC_unit_operation}, some AC units  follow a cycling mode when they are in use whereas other do not. Similar to the observations shown in ~\cite{arjunan2021operational}, such cycling behaviour is attributed to compressor-based AC units.

\begin{figure*}
    \centering
    \includegraphics[scale=0.5]{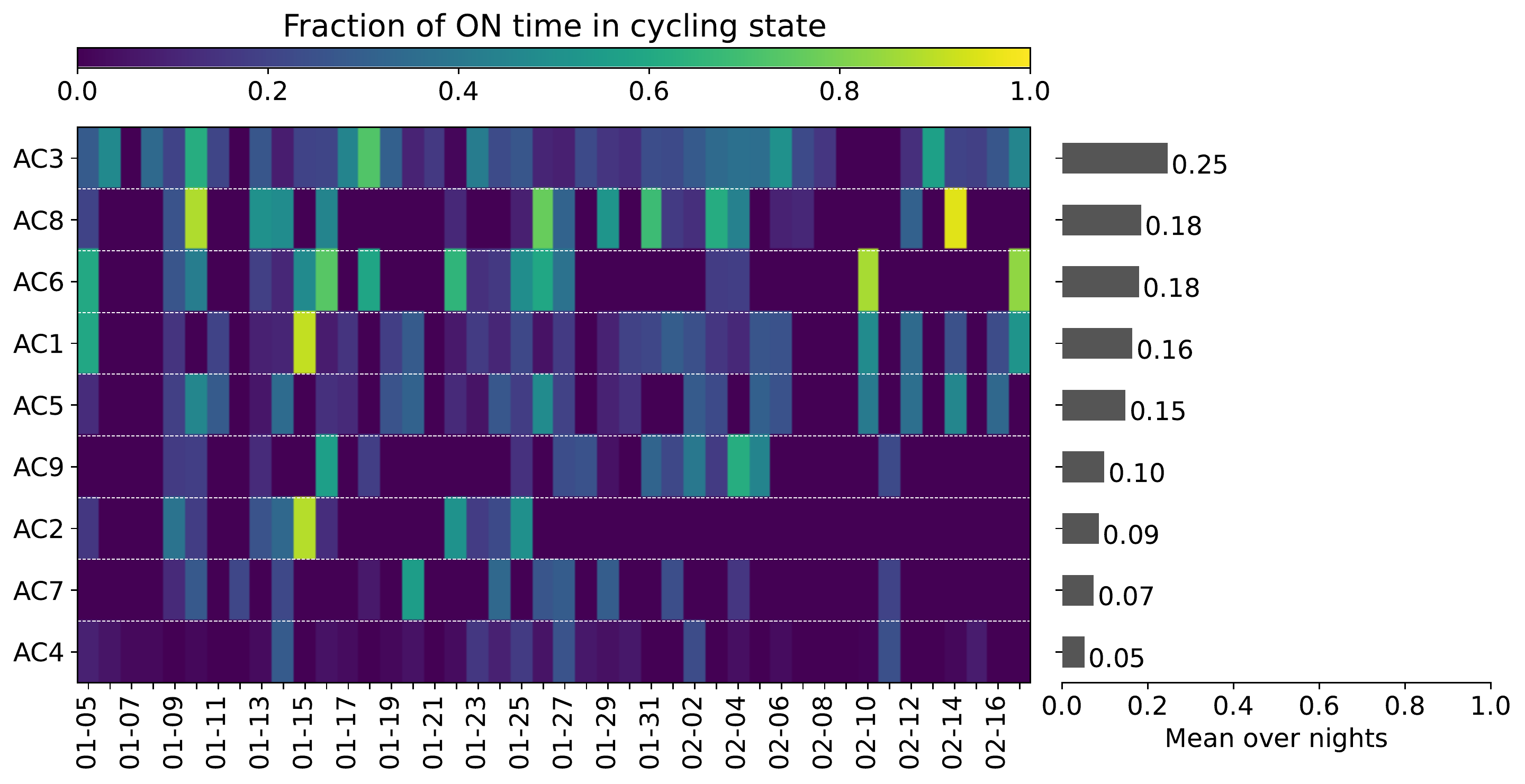}
    \caption{Characterization of nine window/split AC units in our study. \textit{Left:} the fraction of `on' time spent in the `cycling' state for each AC (Y-axis) on each night (X-axis) in our study. While some AC units spend most of the night cycling, others do not cycle at all, potentially indicating different AC types or differences in user set temperatures. \textit{Right:} the average cycling fraction over all nights for each AC. }
    \label{figure:Characterization}
\end{figure*}

Figure~\ref{figure:Characterization} shows the distribution of AC usage by nine occupants across the entire data collection period. We can observe that AC3 is in the cycling state for 25\% of night time. Whereas AC4 is in the cycling state only for 0.5\% of the night time. This variation of AC usage per night could potentially indicate the differences in AC types, usage, and/or set-temperature preferences of the occupants. Further analysis over longer duration is required to identify specific AC type and the occupant's behaviour. 

\subsection{Imaging for maximum information gain}
One of the main advantages of using thermal imaging is that it offers a non-contact method for scanning a large area in a short period. However, there can be certain times of day during which the infrared images that are captured may not yield maximum information gain, which also depends on the type of application for which the images are being used. One such instance is during intense rainfall. During intense rainfall, the infrared images are blurry and do not capture the surface temperature of the buildings. Besides, intense rainfall, as shown in Figure \ref{figure:compare_with_outdoor_temperature} the effect of solar radiation and outdoor temperature has a significant effect on the analysis for characterization of AC usage patterns. The operation pattern from the window AC units can be clearly and accurately detected when the effect of solar radiation is minimal especially between 8 pm and 10 am. This time interval will be suitable for applications involving characterizing the AC system such as its efficiency or detection of faults in its operation. Also, for the AC unit characterization studies, the frequency of data collection should be higher than the frequency of operation of the AC unit to accurately capture the changes in the duty cycle and on/off states.

\section{Conclusion}
Unlike the previous studies on urban scale infrared thermography such as the one by \cite{dobler2021urban} showing the possibility for identification of heat sources in the urban settlement, or on estimation of heat flux by \cite{richters2009analysis} and \cite{ sham2012verification}, this work is a unique demonstration of extraction of the operational pattern of the HVAC system in an educational building using the longitudinal thermal images. To achieve this an IR observatory was installed on the rooftop and operated for a few months.

The temperature data from the thermal camera was first verified against the temperature measured using the surface contact sensors. It is observed that even though there exist differences in the absolute values of the temperature, the de-trended temperature pattern from the thermal images matches that of the temperature recorded using contact sensors. It is important to verify the accuracy in the similarity of the time series pattern because the main focus here is to extract the operational pattern of the HVAC system. However, for studies involving urban heat island (\cite{miguel2022b}) it would be required to re-calibrate the thermal camera constants listed in Table 2 and the effect of boundary conditions needs to be studied. 

Subsequently, a complete pipeline for processing the thermal images, followed by extraction and analysis of temperature time series is demonstrated. In the work by \cite{arjunan2021operational}, the operation of single type of air-conditioning unit in residential buildings was identified. While in this work, the educational building has two different air conditioning systems. It is shown that based on the type of HVAC system, the time series pattern analysis has to be performed either in the time domain or frequency domain.

It is demonstrated that, by observing the changes in the surface temperature profile for several days of the window and wall, the operation pattern of the centralized HVAC system can be deduced. The heatmaps shown in Figure 6 compare the time series analysis of indoor dry bulb temperature against the temperature of the window and wall from the thermal images. If we observe the changes over a few days, a clear pattern corresponding to switching ‘on’ and ‘off’ of the water-cooled HVAC system can be identified. 

For, the individual window AC units, a comparative analysis is performed in the temporal and frequency domain and is shown in Figure 9. The duty cycles of the condenser units identified using wavelet transform are indicative of the ‘on’ and ‘off’ state of the unit. The level of accuracy in the prediction of the ‘on’ and ‘off’ state is shown in Figure 12. The advantage of using wavelet transform is that it is possible to identify not just the duration of operation but also the frequency of the duty cycle of the condenser units.  Studying the changes in the frequency of the duty cycle has been shown to be useful for the evaluation of the AC unit efficiency and also to detect faults (\cite{rafati2022fault, rogers2019review}). In addition to this, insights on usage patterns can be indicative of human behavior as these units are operated by occupants themselves (\cite{yasue2013modeling, xia2019study, aqilah2021analysis}). To summarize, these are the main research contributions:
\begin{itemize}
  \item A longitudinal thermal imaging technique at urban scale is demonstrated for studying the HVAC operational pattern in an educational building.
  \item The operational pattern of a water-cooled HVAC system can be detected by analysing the window and wall temperature simultaneously over a duration of a few days.
  \item A wavelet transform of the thermal signature of the condenser unit of the air-cooled HVAC system is used for extraction of its operational pattern.
\end{itemize}

Inference from the analysis of longitudinal thermal images can serve as an alternative data-driven means to conduct energy audits (\cite{IEA2021cooling}). One of the main advantages of this type of analysis compared to conventional energy audit is that it does not require the deployment of a large array of temperature sensors or energy meters (\cite{zhou2021comparison, yun2011field}) in the building space. There are possibilities to improve the accuracy in prediction of the operational pattern and user behaviour by considering the effect of solar radiance and outdoor temperature and it is aimed to address these in future work.

\bibliographystyle{elsarticle-num-names}

\bibliography{cas-refs}


\end{document}